\documentclass[10pt,twocolumn,letterpaper]{article}

\usepackage{cvpr}              

\newcommand{\debiaslens}{\textsc{DeBiasLens}\xspace}

\usepackage[most]{tcolorbox}
\newcounter{findingcounter}
\newtcolorbox{findingbox}{
  colback=black!5,        
  colframe=black!15,      
  boxrule=0.3pt,
  arc=2pt,
  left=6pt,
  right=6pt,
  top=4pt,
  bottom=4pt,
  before skip=6pt,
  after skip=6pt,
}

\definecolor{cvprblue}{rgb}{0.21,0.49,0.74}
\usepackage[pagebackref,breaklinks,colorlinks,allcolors=cvprblue]{hyperref}

\usepackage{url}
\usepackage{booktabs} 
\usepackage{multirow}
\usepackage{graphicx}
\usepackage[table]{xcolor}
\usepackage[skip=6pt]{caption}
\usepackage[margin=1in]{geometry}
\usepackage{adjustbox}
\usepackage{tcolorbox}
\usepackage{titletoc}
\usepackage{lipsum}

\usepackage{tikz}
\newcommand{\tikzxmark}{%
\tikz[scale=0.23] {
    \draw[line width=0.7,line cap=round] (0,0) to [bend left=6] (1,1);
    \draw[line width=0.7,line cap=round] (0.2,0.95) to [bend right=3] (0.8,0.05);
}}

\definecolor{oursblue}{RGB}{230,245,255}
\definecolor{oursred}{RGB}{255,220,220}
\definecolor{ourspurple}{RGB}{235,220,250}

\def\paperID{} 
\def\confName{CVPR}
\def\confYear{2025}

\title{Interpretable Debiasing of Vision-Language Models for Social Fairness}

\author{
Na Min An$^{1}$ \quad
Yoonna Jang$^{2}$ \quad
Yusuke Hirota$^{3}$ \\
Ryo Hachiuma$^{3}$ \quad
Isabelle Augenstein$^{2}$ \quad
Hyunjung Shim$^{1}$ \\
$^{1}$ KAIST AI \quad 
$^{2}$ University of Copenhagen \quad 
$^{3}$ NVIDIA \\
\tt\small{\{naminan\}@kaist.ac.kr}
}

\begin{document}
\twocolumn[{
\maketitle
\begin{center}
    \makebox[\textwidth][c]{%
        \includegraphics[width=\textwidth]{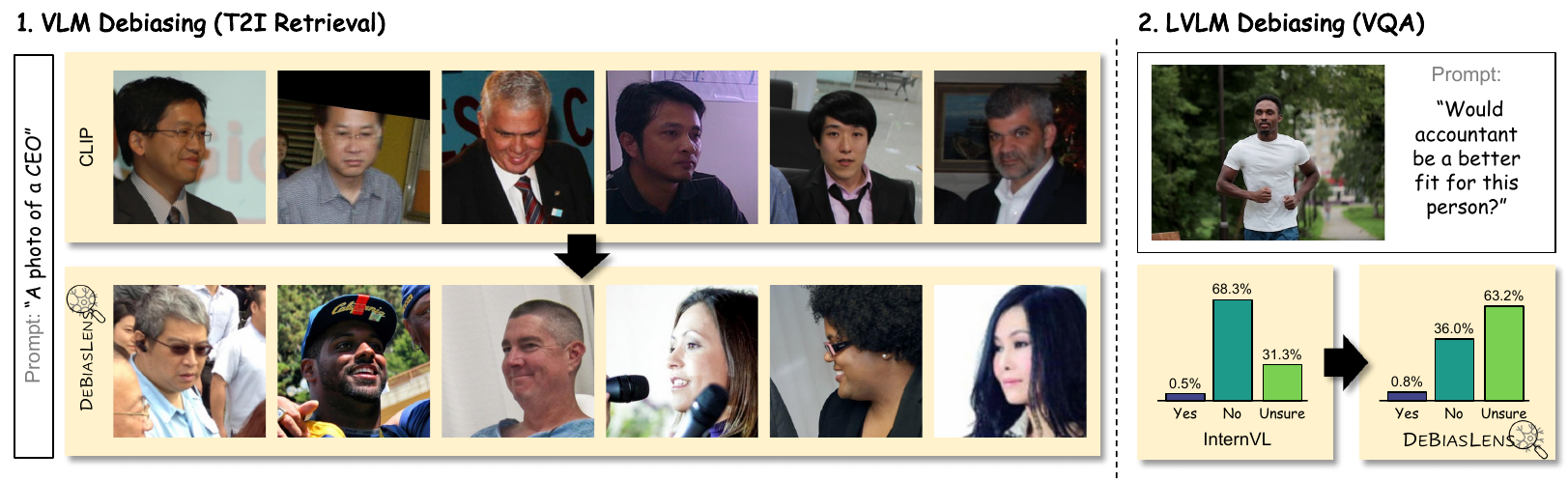}
    }
    \vspace{-3mm}
    \captionof{figure}{
        \textbf{Social bias mitigation in VLMs.} While existing models retrieve image distribution of skewed demographics or answer definitively on ambiguous image-text pairs, our \debiaslens alleviates social biases across both image and text modalities.
    }
    \label{fig:main}
\end{center}

}]
\begin{abstract}
The rapid advancement of Vision-Language models (VLMs) has raised growing concerns that their black-box reasoning processes could lead to unintended forms of social bias. Current debiasing approaches focus on mitigating surface-level bias signals through post-hoc learning or test-time algorithms, while leaving the internal dynamics of the model largely unexplored. In this work, we introduce an interpretable, model-agnostic bias mitigation framework, \textsc{DeBiasLens}, that localizes social attribute neurons in VLMs through sparse autoencoders (SAEs) applied to multimodal encoders. Building upon the disentanglement ability of SAEs, we train them on facial image or caption datasets without corresponding social attribute labels to uncover neurons highly responsive to specific demographics, including those that are underrepresented. By selectively deactivating the social neurons most strongly tied to bias for each group, we effectively mitigate socially biased behaviors of VLMs without degrading their semantic knowledge. Our research lays the groundwork for future auditing tools, prioritizing social fairness in emerging real-world AI systems.
\end{abstract}    
\section{Introduction}
\label{sec:intro}

Recent advancements in foundational Vision–Language Models (VLMs) (\eg, CLIP~\citep{radford2021learning})~\citep{zhai2023sigmoid} and large VLMs (LVLMs) (\eg, InternVL~\citep{chen2024internvl}) integrated with Large Language Model (LLM)~\citep{liu2023visual, yang2025qwen3, hurst2024gpt} have shown possibilities for deployment in high-impact applications, such as assistive technologies~\cite{kim2025multi,kang-etal-2025-sightation}. 
However, the rapid adoption of these models also increases concerns around fairness and social responsibility~\citep{ryan2021artificial,birhane2021multimodal,park2025alignedstereotypicalhiddeninfluence}. Since VLMs and LVLMs are trained on large-scale multimodal datasets, they may inherit and even amplify societal biases learned from the training data~\citep{seth2023dear,lee2023survey,wang2021gender,chuang2023debiasing,hamidieh2024identifying,howard2025uncovering,kim2025tom,wu2025model,khan2025investigating}. For example, CLIP retrieves images skewed to certain demographics (\eg, male) when prompted with seemingly neutral descriptions (\eg, ``A photo of a CEO'') in text-to-image (T2I) retrieval, and InternVL provides definitive answers in visual question answering (VQA) tasks even under ambiguous contexts (see Figure~\ref{fig:main}). These biased outputs not only mischaracterize reality but also reinforce discriminatory visual norms at scale, posing serious harm to users and underrepresented groups or cultures~\cite{um2023fair,kim2025tom}. Ultimately, addressing and mitigating such biases in a reliable and principled manner has become a crucial research challenge.

Since it is costly to pre-train VLMs and LVLMs from scratch, most existing debiasing methods rely on post-hoc learning algorithms (\eg, fine-tuning, prompt tuning) and test-time debiasing approaches (\eg, pruning, prompt engineering)~\citep{hirota2025saner, girrbach2025revealing, chuang2023debiasing, gerych2024bendvlm}. For instance, \citet{hirota2025saner} proposed a debiasing approach that fine-tunes CLIP on the gender-balanced datasets by adding a residual layer to mitigate the harmful gender-occupation associations. \citet{girrbach2025revealing} found that the LoRA fine-tuning~\citep{hu2021lora} is the most promising approach when debiasing LVLMs, which shows the lowest trade-off between general and bias mitigation performance.

However, we argue that these debiasing methods suffer from a critical limitation: They overlook the model's \emph{internal dynamics}. Consequently, they merely alleviate surface-level symptoms of biased behavior without modifying the underlying internal representations through which bias propagates~\citep{gerych2024bendvlm,jiang2024modscan}. This lack of interpretability makes it difficult to identify and target the components responsible for encoding social biases, subsequently hindering precise debiasing and often degrading the model's original representations. For instance, although the model weight pruning approach aims to identify parameters that strongly influence bias mitigation while minimally affecting overall performance, it paradoxically achieves bias reduction at the cost of substantially compromising the model’s general capability~\citep{girrbach2025revealing}. This is likely since individual model neurons (linked to the weights) often encode \emph{polysemantic} concepts~\cite{pach2025sparse}, simultaneously affecting both bias and general model capability.

To address this research gap, we propose \debiaslens, a debiasing method that identifies bias-related social neurons and applies targeted, minimal interventions to effectively mitigate bias. Our method is inspired by recent work leveraging sparse autoencoders (SAEs) for feature disentanglement in LVLM~\cite{pach2025sparse}, which enables the extraction of semantically disentangled, \emph{monosemantic} neurons~\citep{anthropic-2023,yan2025multi}. Since social bias emerges from strong, consistent correlations associated with specific demographics, we hypothesize that, when trained with appropriate datasets and settings, the SAE can encode ``\emph{social neurons}'' that enable social bias attribute-specific (\eg, gender, age, race) targeted mitigation of biased model behavior. Crucially, our method provides transparent, neuron-level debiasing by modulating these social neurons, enabling effective mitigation while maintaining the model’s original capabilities across diverse multimodal tasks. Specifically, we achieve a 9--16\% reduction in Max Skew for CLIP image retrieval and a 40--50\% decrease in gender disproportion for InternVL2, all while preserving the original performance across general VLM reasoning tasks. The simplicity of our approach makes it readily applicable to both VLMs and LVLMs (Figure~\ref{fig:main}). 

In sum, our contributions are as follows:
\begin{itemize}
    \item The first interpretable debiasing mitigation framework applicable for VLM and LVLMs.
    \item Effective bias mitigation strategy, preserving overall performance in general VLM tasks.
    \item Informative guide on how to utilize SAE for developing bias-aware multimodal systems.
\end{itemize}

\section{Related Work}
\label{sec:rel-work}

Our work bridges the fields of bias mitigation in VLMs and mechanistic interpretability.

\begin{figure*}[!t]
    \centering
    \includegraphics[width=\textwidth]{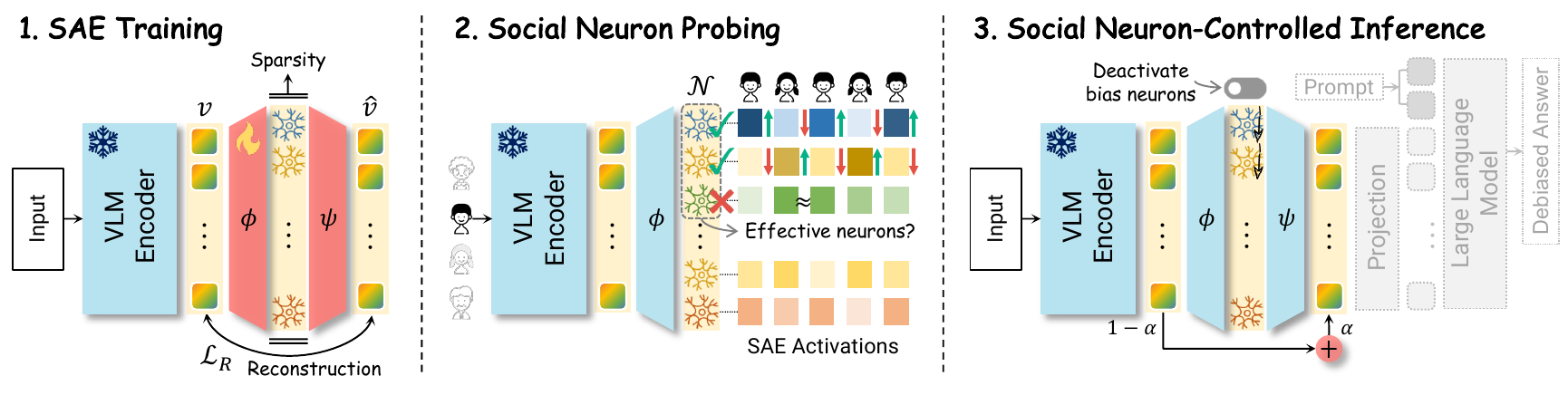}
    \caption{\textbf{Overview of our interpretable VLM debiasing framework.} \debiaslens consists of three stages: (1) SAE is trained on top of the last layer of the VLM image/text encoder (Section~\ref{sec:method1}). (2) The social neurons are identified based on the consistency and specificity of SAE activations across data (Section~\ref{sec:method2}). (3) The selected neurons are activated to generate debiased features, weighted summed with original features for further usage across downstream tasks (Section~\ref{sec:method3}).}
    \label{fig:framework}
    \vspace{-1em}
\end{figure*}

\paragraph{Bias Mitigation in Vision-Language Models}

Mitigating biases in VLMs has been mostly conducted via post-hoc learning methods, aiming to preserve the original image or text embedding representations. Model pruning and fine-tuning are effective in reducing biases, but simultaneously, struggle with catastrophic forgetting~\cite{gerych2024bendvlm,girrbach2025revealing}. Hence, several methods that do not need full fine-tuning have been introduced: \citet{berg2022prompt} debias VLMs by prepending learnable embeddings to text queries trained with an adversarial classifier, leaving both the text and image encoders frozen. \citet{seth2023dear} learns additive residual image representations for neutralization. \citet{chuang2023debiasing} propose a closed-form method to debias VLMs by projecting out biased directions from text embeddings. \citet{kong2024mitigating} use fair retrieval subsets using off-the-shelf social attribute classifier or VLM. \citet{gerych2024bendvlm} present a debiasing operation tailored to each input. \citet{hirota2025saner} debiases text embeddings by training a residual layer without using social attribute annotations.

While the aforementioned works focus on mitigating either image-only or text-only representations, several other studies probe into both modalities: \citet{jung2024unified} combine feature pruning and low confidence imputation for effective bias reduction. \citet{weng2024images} apply causal mediation analysis~\citep{vig2020investigating} to identify the direct and indirect interventions on model bias, revealing that image features are the primary contributors to bias in VLMs. \citet{jiang2024modscan} utilize specific prompt prefixes in the language or vision input to reduce stereotypical biases in VLMs. \citet{zhang2025joint} propose a framework that jointly aligns and removes biases from both modalities to achieve fairer representations without sacrificing image-text representation alignment~\citep{an2024i0t}. Our work is similarly applicable for image and/or text modalities.

\paragraph{Mechanistic Interpretability}

The goal in the domain of mechanistic interpretability (MI) is to reverse-engineer neural networks, translating their learned parameters into human-interpretable algorithms~\citep{Nanda_2021b, geva2021transformer} to explain how and why they arrive at their answers~\citep{saphra2024mechanistic,Hastings-Woodhouse_2024}. MI can also be applied to VLMs to uncover their internal mechanisms: \citet{cao2020behind} use probing~\citep{alain2016understanding,hewitt2019structural,bai2024describe} technique to reveal how multimodal pre-training shapes attention patterns. \citet{salin2022vision} \textit{probe} the reliance of VLMs on biased cues for diverse multimodal tasks. \citet{palit2023towards} and \citet{golovanevsky-etal-2025-vlms} apply activation patching (\ie causal tracing)~\citep{vig2020investigating,meng2022locating,zhang2025cross} by selectively controlling specific internal activations to identify critical layers and components in VLMs. Logit lens~\citep{nostalgebraist_2020,belrose2023eliciting} technique is used to analyze how VLM processes visual tokens for object identification~\citep{neo2024towards} and visual hallucination detection~\citep{jiang2024interpreting,phukan2025beyond,jiang2025devils}.

A widely adopted technique in MI leverages Sparse AutoEncoders (SAEs) to project original features into a sparse, interpretable representation via a high-dimensional expansion layer~\citep{arora2018linear,olah2020zoom,nabeshima2024matryoshka,peng2025use}. Although there exist works that attempt to find task/domain-specific interpretable attention heads and neurons within the pre-trained VLMs~\citep{golovanevsky2025vlms,namazifard2025isolating,huo2024mmneuron,huang2024miner}, more fine-grained semantic features within each modality are entangled in these neurons~\citep{pach2025sparse,kim2025interpreting}. While SAEs have been applied to show effectiveness for bias mitigation in text-to-image (T2I) diffusion models~\citep{surkov2024one,shi2025dissecting,tian2025sparse}, it has yet to be explored for encoder-based and image-text-to-text (IT2T) VLMs for the bias mitigation task. One of the reasons may be the difficulty of finding how to \emph{localize} and \emph{regulate} SAE social neurons. This paper addresses this challenge by introducing an interpretable framework that identifies and selectively modulates neurons responsible for encoding social biases.
\section{Methodology}
\label{sec:method}

Our goal is to mitigate social bias in pretrained vision–language models (VLMs) without modifying model weights or retraining, while preserving general performance and ensuring interpretability. To this end, we introduce \debiaslens, a novel bias mitigation framework (Figure~\ref{fig:framework}), which enables interpretable debiasing that transparently disentangles and regulates social neurons within the models' internal representations.

\subsection{SAE Training}\label{sec:method1}
The first step transforms the entangled feature space of the VLM encoder (post-residual) into a sparse, interpretable latent space using SAE. Inspired by recent findings in mechanistic interpretability with monosemantic SAE neurons \citep{gao2024scaling,pach2025sparse}, we attach an SAE layer ($\phi(\cdot)$) to the last layer of the pretrained VLM encoder (either vision, text, or both). Specifically, the SAE is trained to decompose the original feature $\mathbf{v} \in \mathbb{R}^d$ into a sparse activation vector $\phi(\mathbf{v}) \in \mathbb{R}^\omega, \omega \geq d$ and reconstruct it as $\hat{\mathbf{v}}$. Mathematically, $\phi(\mathbf{v}) = \sigma(\mathbf{W}_{\rm enc}^\top (\mathbf{v}-\mathbf{b}_1)), \hat{\mathbf{v}} = \psi(\phi(\mathbf{v})) = \mathbf{W}_{\rm dec}^\top \phi(\mathbf{v}) + \mathbf{b}_2$. Here, $\sigma$ is a nonlinearity function (\eg, ReLU), and $\mathbf{W}_{\rm enc}, \mathbf{W}_{\rm dec}$, and $\mathbf{b_i}$ are learnable parameters optimized by minimizing reconstruction error while enforcing sparsity with weight ($\lambda$ in Eq.~\ref{eq:2}). We adopt Matryoshka SAEs \cite{bussmann2025learning, nabeshima2024matryoshka} to form the reconstruction objective into a multi-scale loss, ensuring that the model learns accurate reconstructions at different levels of sparsity. If $\mathcal{M}$ is a set consisting of different depths of SAE layers, and $\phi_{1:m}(\mathbf{v})$ denotes the top-$m$ active neurons, the main objective $\mathcal{L}_1(\mathbf{v})$ is:

\vspace{-1em}
\begin{equation}
\begin{split}
    \mathcal{L}_1(\mathbf{v}) & = \mathcal{L}_R(\mathbf{v}) + \lambda \|\phi(\mathbf{v})\|_1, \label{eq:2} \\
    \mathcal{L}_R(\mathbf{v}) & = \sum_{m \in \mathcal{M}} \left\|\mathbf{v} - \mathbf{W}_{\rm dec}^\top \phi_{1:m}(\mathbf{v})\right\|_2^2.
\end{split}
\end{equation}

\begin{figure}[!t]
    \centering
    \includegraphics[width=\linewidth]{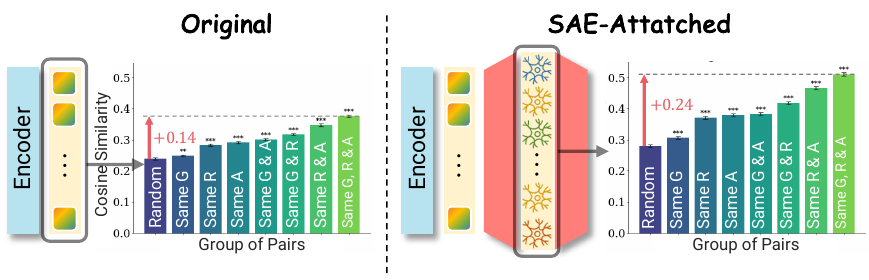}
    \caption{\textbf{Comparison between similarity trend of facial image pairs for original and SAE-attached CLIP.} The difference between the cosine similarity of random and social attribute-overlapping image pairs (G: gender, R: race, A: age) becomes more pronounced when our SAE is attached, indicating that the SAE can capture latent bias-sensitive features.}
    \label{fig:motivation}
    \vspace{-1em}
\end{figure}

We also implement auxiliary loss~\citep{gao2024scaling} to utilize the top-$m_{\mathrm{aux}}$ dead latents for modeling reconstruction error. Hence, for $\mathbf{e} = \mathbf{v} - \hat{\mathbf{v}}$, and $\hat{\mathbf{e}} = \mathbf{W}_{\rm dec}\phi_{1:m_{\mathrm{aux}}}(\mathbf{v})$, the final loss is defined as: $\mathcal{L}(v) = \mathcal{L}_{1}(\mathbf{v}) + \beta  \|\mathbf{e} - \hat{\mathbf{e}}\|_2^2.$

\noindent\textbf{Why SAE?} Finding interpretable SAE neurons allows direct modulation of the internal model structure without hurting the original architecture by retraining or pruning. In essence, the SAE serves as an analytical lens that reveals and disentangles the internal bias structures within frozen VLMs. Leaving the original model unchanged, only the SAE is trained on any type of facial-related image/caption datasets without social attribute labels.

\subsection{Social Neuron Probing}\label{sec:method2}
Based on the observation that SAE neurons can implicitly capture social attributes (Figure~\ref{fig:motivation}) even without explicit supervision from demographic labels (\eg, gender, age, or race), we hypothesize that the \emph{social neurons} contributing to a model's bias exhibit differential SAE activation patterns across data groups representing different social attributes. To locate these social neurons, we first quantify the effectiveness score of an SAE neuron $j$ within a social attribute group $g$ (\eg, female for gender), we use the following effectiveness criterion: $\sum_{i=1}^{S_g} \mathbb{I}(\mathbf{x}_{i, j}^{(g)} \neq 0) \ge \lfloor \tau \cdot S_g \rfloor$.

Here, $\mathbf{x}_{i, j}^{(g)}$ is the SAE activation of neuron $j$ for sample $i$ in group $g$, and $\mathbb{I}(\cdot)$ is the indicator function. The effective neurons are automatically selected if their SAE activations are non-zero for at least $\tau$ proportion of the samples for each group with the size of $S_g$. 

This process results in the set of automatically selected effective neurons, $\mathcal{E}_g$. Subsequently, we identify the social attribute group-specific neurons $\mathcal{N}_g$ by calculating the set difference $\mathcal{N}_g = \mathcal{E}_g \setminus \mathcal{U}_{\neg g}$, where $\mathcal{U}_{\neg g}$ is the union of effective neurons from all other social attribute groups $h \neq g$:

\vspace{-1.5em}
$$\mathcal{N}_g = \mathcal{E}_g \setminus \left( \bigcup_{h \in G, h \neq g} \mathcal{E}_h \right)$$
\vspace{-0.5em}

The set $\mathcal{N}_g$ thus comprises neurons that are activated almost universally within group $g$ but not within any other group, making them strong candidates for encoding the group's specific features. These candidate neurons are ranked by their mean activation value $\bar{s}_j$ within group $g$ to prioritize those with the strongest within-group signal: $\bar{\mathbf{s}}_j = \frac{1}{S_g} \sum_{i=1}^{S_g} \mathbf{x}_{i, j}^{(g)}, \text{for } j \in \mathcal{N}_g$

Finally, we select \emph{social neurons} for each group as the neuron $j \in \mathcal{N}_g$ that exhibits the highest mean activation value: $j_{g}^{*} = \arg\max_{j \in \mathcal{N}_g} \left( \bar{\mathbf{s}}_j \right)$. We store the selected social neurons (\eg, all female and male neurons selected for `gender' attributes) in $\mathcal{Z}_{\text{B}}$.

\subsection{Social Neuron-Modulated Inference}\label{sec:method3}

During the inference phase, the SAE activations corresponding to the social neurons are neutralized (or deactivated) by setting their activation values in the latent vector $\mathbf{z}$ to zero. Specifically, for all neuron indices $j \in \mathcal{Z}_{\text{B}}$, the corresponding component of $\mathbf{z}'$ is set as:

\vspace{-0.5em}
$$
\mathbf{z}'[j] =
\begin{cases}
    \gamma & \text{if } j \in \mathcal{Z}_{\text{B}} \\
    \mathbf{z}[j] & \text{otherwise}
\end{cases}
$$

Following this targeted deactivation ($\gamma$ set to 0 for most cases but allowed to take negative values depending on the desired strength of the deactivation), the modified latent vector $\mathbf{z}'$ is passed through the trained SAE decoder to generate the bias-free reconstructed feature: $\hat{\mathbf{v}} = \psi(\mathbf{z}')$. Since the SAE is specifically trained to capture social neurons, the reconstructed feature $\hat{\mathbf{v}}$ might be shifted towards the latent space of the SAE training data distribution. To proportionally utilize the reconstructed feature provided by the SAE while preserving the original feature's information, we use a weighted sum: $\mathbf{v}' = \alpha \hat{\mathbf{v}} + (1 - \alpha)  \mathbf{v}$, where $\alpha \in [0, 1]$ is weight proportion. The resulting mixed vector $\mathbf{v}'$ replaces the original hidden state in the subsequent operations of the transformer block, mediating the effect of potentially biased feature components detected and deactivated via SAE. The novelty of our method lies in leveraging SAE to isolate bias as recurring internal features within a subset of interpretable neurons for targeted bias mitigation.
\section{Experiments}
\label{sec:dataset}

This section lays out implementation details of our methodology and evaluation on the effectiveness of our \debiaslens. Details are in the Supp. \ref{app:training_details}, \ref{app:data_details} and \ref{app:add_results}.

\subsection{Experimental Details}

\paragraph{SAE Training datasets} To examine the effect of training data distribution on the reconstructed latent space, we train SAE using various data configurations: 
CelebA~\citep{liu2018large}, Cocogender images~\citep{tang2021mitigating}, and
FairFace~\citep{karkkainen2019fairface} for the image encoder and Cocogender captions (Cocogendertxt)~\citep{tang2021mitigating} and Bias in Bios~\citep{de2019bias} for the text encoder. Note that all these datasets have `gender' as a bias attribute label, and the FairFace dataset also has `age' and `race' labels.

\paragraph{Evaluation datasets} To assess the social bias level of encoder-based VLMs (CLIP variants~\citep{radford2021learning}, following previous VLM-related works~\citep{berg2022prompt, chuang2023debiasing, gerych2024bendvlm, hirota2025saner}) and encoder-decoder-based LVLMs (LLaVA-1.5-7B~\citep{liu2023visual} and InternVL2-8B~\citep{chen2024internvl}, known to show the strongest gender bias tendency~\citep{girrbach2025revealing} among LVLMs), we evaluate the models on the T2I retrieval and VQA tasks, respectively. When evaluating the T2I retrieval, we use 10,954 cropped facial datasets in FairFace ~\citep{karkkainen2019fairface}, following \citep{hirota2025saner,seth2023dear}. The social bias attribute labels in these datasets are gender, age, and race. The text prompts used as inputs for T2I retrieval consist of 29 adjectives, 41 occupations, and 33 activities (spanning 12, 7, and 6 unique templates)~\citep{hirota2025saner}. We exclude the age groups of ``0-2'' and ``more than 70'' for the occupation category. We also test 25 stereotype text prompts~\citep{gerych2024bendvlm}.

When evaluating LVLMs, we use two recent benchmarks: VLAGenderBias (VLA)~\citep{girrbach2025revealing} and SBBench~\citep{narnaware2025sb}. The former consists of 5k facial images with gender bias attributes collected from various data sources~\citep{karkkainen2019fairface,seth2023dear,schumann2021step,garcia2023uncurated}, and the latter benchmark consists of 14.6k images labeled with nine social attributes\footnote{Due to the sample insufficiency for certain social attributes (\eg, physical appearance), we test perceived gender and age attributes.} pooled from web search. The sources of text prompts are pooled from various sources~\citep{zhao2018gender,kurita2019measuring,jiechieu2021skills} for VLA (prompts: occupation, sentiment, and skills) and BBQ~\citep{parrish2021bbq} for SBBench. All these datasets are distributed fairly across pre-defined social attributes (\eg, gender). We use ImageNette~\citep{Howard_Imagenette_2019} and VLMEvalKit MME~\citep{chaoyou2023mme}-measured with the sum of perception and reasoning scores, MMMU-dev~\citep{yue2024mmmu}, and Seed-Bench-2~\citep{li2024seed}) for general performance evaluation~\citep{duan2024vlmevalkit}.

\paragraph{Evaluation metrics} Following previous works~\citep{chuang2023debiasing, berg2022prompt, gerych2024bendvlm,  jung2024unified, hirota2024picture}, we use Max Skew to quantify how much the distribution of retrieved images approximates a uniform distribution across different demographics per bias attribute. For benchmarking LVLMs, the bias is measured in two ways: (1) The proportion of statistically different answers of ``yes'' between male and female-labeled data (denoted as gender disproportion rate)~\citep{girrbach2025revealing}, and (2) The accuracy of correctly answering bias probing questions, such as opting for ``cannot be determined,'' given the image-text pair with no definite correct answers~\citep{narnaware2025sb}.

\paragraph{Comparison methods}
We compare our approach applied to VLM with the following methods: \textbf{Prompt}~\citep{berg2022prompt} learns prefix embeddings for text queries using a combined adversarial and contrastive learning objective. \textbf{Projection}~\citep{chuang2023debiasing} uses a closed-form projection matrix to remove biased directions from the text embeddings without additional data or training. \textbf{Bend-VLM}~\citep{gerych2024bendvlm} is a nonlinear, fine-tuning-free approach that debiases VLM embeddings by customizing the debiasing process for each input using spurious and augmented prompts. \textbf{SANER}~\citep{hirota2025saner} neutralizes text features by erasing attribute information only from attribute-neutral text inputs. We use the variation of \textbf{MMNeuron}~\citep{huo2024mmneuron}, which identifies bias attribute-specific neurons within the pre-trained layers of VLM encoders.

The comparison methods for LVLMs are as follows~\citep{girrbach2025revealing}: \textbf{Full Fine-Tuning} and \textbf{LoRA Fine-Tuning} optimize all model parameters and low-rank adapters (LoRAs~\citep{hu2021lora}). \textbf{Pruning} method identifies and prunes the parameters that are effective in mitigating bias but show less influence on general loss performance. \textbf{Prompt Tuning} trains embeddings of a soft prompt prefix that can be transferable across prompt variations for bias reduction. \textbf{Prompt Engineering} inputs debiasing instructions, such as ``Please, be mindful that people should not be judged based on their race, gender,
age, or other physical characteristics,'' during the test stage.

\begin{table}[!t]
\centering
\caption{\textbf{Max Skew@1000 (scaled by 100) results on FairFace dataset using diverse prompts for gender bias evaluation}. Note $\dag$ represents reproduced results (T and I indicate SAE attached to text and image encoder). Our \debiaslens attains comparable performance with SoTA VLM debiasing methods without using labels during training (but required during probing), with interpretable inference components.}
\resizebox{\linewidth}{!}{
\begin{tabular}{lccccc}
\toprule

\multicolumn{1}{c}{\multirow{2}*{\textbf{Methods}}} & \multirow{2}*{\textbf{Interpretable?}} & \multicolumn{4}{c}{\textbf{Max Skew ($\downarrow$)}}\\
\cmidrule(lr){3-6}
 &  & \textbf{Adj} & \textbf{Occup} & \textbf{Act} & \textbf{Ster} \\
\midrule
CLIP (ViT-B/16)~\citep{radford2021learning} & $-$ & 22.9 & 33.7 & 19.5 & 33.8 \\
CLIP (ViT-B/16)$\dag$ & $-$ & 21.9 & 33.5 & 19.8 & 32.5 \\ 

Prompt~\cite{berg2022prompt} & \tikzxmark & 12.3 & 29.9 & 20.0 & -\\
Prompt$\dag$ & \tikzxmark & 11.9 & 29.8 & 19.3 & 28.7 \\

Projection~\cite{chuang2023debiasing} & \tikzxmark & 15.4 & 37.4 & 15.0 & 52.0 \\

Bend-VLM$\dag$~\cite{gerych2024bendvlm} & \tikzxmark & 10.8 & \textbf{10.2} & \underline{9.8} & \underline{9.1} \\

SANER~\cite{hirota2025saner} & \tikzxmark & \underline{8.9} & \underline{14.5} & \textbf{7.7} & - \\

\rowcolor{gray!15} \debiaslens (I) & \checkmark & 14.2 & 21.5 & 20.0 & 18.3 \\
\rowcolor{gray!15} \debiaslens (T) & \checkmark & \textbf{7.1} & 16.2 & 14.2 & \textbf{8.1} \\
\rowcolor{gray!15} \debiaslens (I+T) & \checkmark & 11.1 & 19.4 & 18.0 & 10.3 \\

\midrule
CLIP (ViT-L/14@336)$\dag$  & $-$ & 19.9 & 31.5 & 23.2 & 30.0 \\

MMNeuron~\cite{chaoyou2023mme}  & \checkmark & 17.3 & 23.5 & 26.0 & 20.3 \\

\rowcolor{gray!15} \debiaslens (I) & \checkmark & \textbf{12.0} & \textbf{20.4} & \textbf{17.2} & \textbf{11.2} \\

\rowcolor{gray!15} \debiaslens (T) & \checkmark & 16.3 & 27.6 & 26.9 & 21.2 \\

\rowcolor{gray!15} \debiaslens (I+T) & \checkmark & \underline{16.2} & \underline{25.1} & \underline{19.9} & 24.2 \\

\bottomrule
\end{tabular}
}
\label{tab:vlm_bias}
\vspace{-1em}
\end{table}

\paragraph{Implementation details} We set the expansion factor of SAE to be 8~\citep{pach2025sparse}, $\tau$ to be 0.9, $\alpha$ to be 0.6 based on the results in the Experiments section (Section~\ref{exp:int}). We use SAE-trained and its neurons probed using FairFace and Cocogendertxt datasets for the image and/or text encoder throughout the experiments if not specified.

\subsection{Debiasing Vision-Language Models}
Applying interpretable \debiaslens to the image and/or text encoder of two widely used CLIP variants~\citep{radford2021learning}, we notice a significant decrease in Max Skew scores (Table~\ref{tab:vlm_bias}), comparable to previous SoTA debiasing methods~\citep{hirota2025saner, berg2022prompt}. Notably, \debiaslens (T) achieves the best bias mitigation results for adjective and stereotype prompts without using attribute labels for training like \citep{hirota2025saner} and augmentation like \citep{gerych2024bendvlm}. The qualitative results (Figure~\ref{fig:main}) also show significant gender balance improvement.

\begin{figure}[!t]
    \centering
    \includegraphics[width=\linewidth]{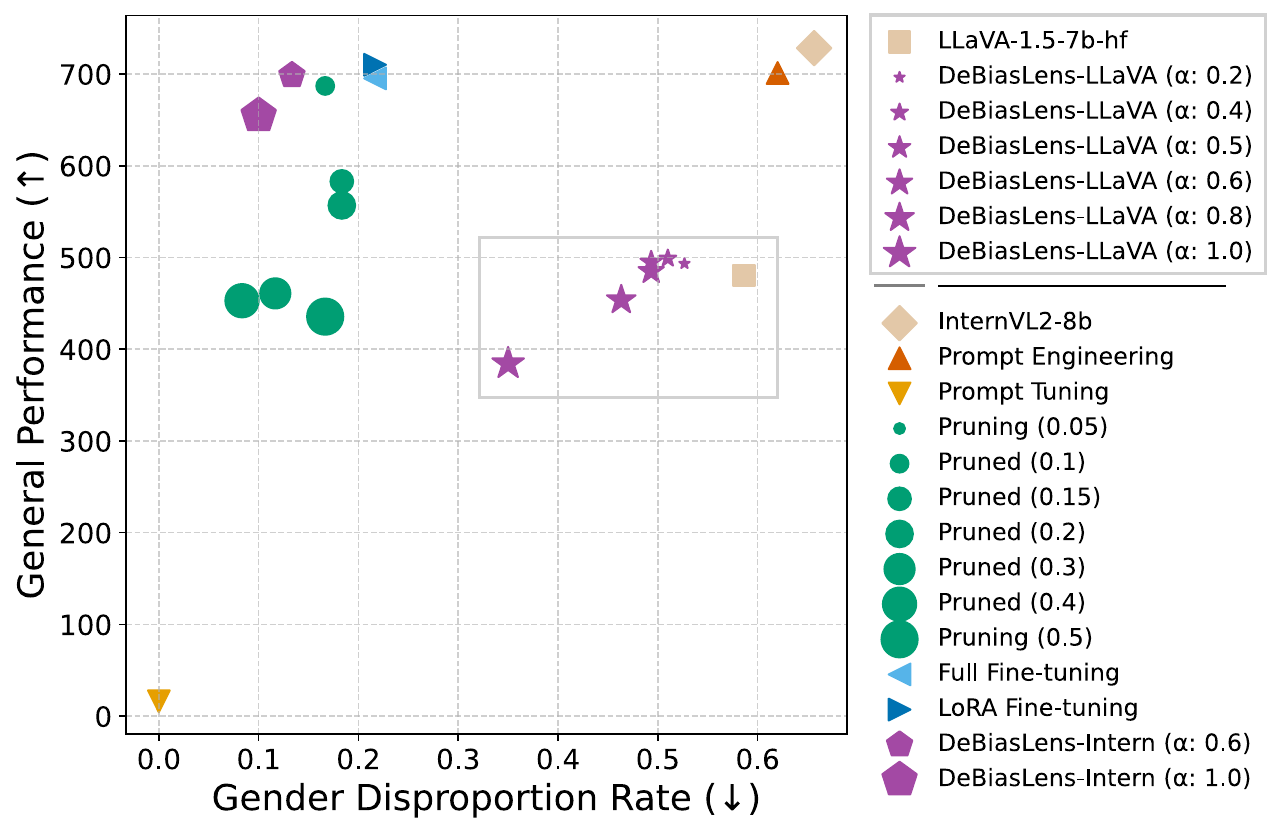}
    \caption{\textbf{Comparison between bias mitigation \textit{vs}. general performance of LVLMs.} Our method achieves the best trade-off among existing approaches ($\leftarrow$, $\uparrow$, the better).}
    \label{fig:vla}
    \vspace{-1.5em}
\end{figure}

Interestingly, while \debiaslens (T) shows better results in CLIP (ViT-B/16) than \debiaslens (I), it is vice versa for CLIP (ViT-L/14@336). This suggests that debiasing the image encoder is more effective for VLM with higher image encoding resolution~\cite{wang2025textspeaksloudervision}, motivating us to explore the image encoder debiasing capability on LVLMs later. Furthermore, we notice the Max Skew scores of \debiaslens (I$+$T), where our debiasing methods are independently applied to both image and text encoders, lie between those that use image-only \debiaslens (I) and text-only \debiaslens (T) VLM encoders. This implies that debiasing each modality partially mitigates bias, but their combined effect is not purely additive.

Figure~\ref{fig:vla} demonstrates our method when applied to LVLMs. We observe that \debiaslens-Intern ($\alpha$: 0.6) shows a reduction of 40--50\% in disproportions that they answer differently across genders, it simultaneously shows a comparatively minimal drop of 4--10 in average general performance (computational cost results in Supp.~\ref{app:add_results}). The raw probability scores of  \debiaslens-Intern across genders per skill/occupation/sentiment-related textual prompts are in Figures~\ref{fig:vla_ex}, \ref{fig:vla_ex2}, and \ref{fig:vla_ex3} in Appendix~\ref{app:add_results}, which illustrate that our method significantly reduces the gender disproportion rate across different types of prompts. For instance, while the original baseline shows a statistically significant difference in prompts such as  ``handle multitasks,''  ``maintain consistency,'' and ``work under pressure,'' our method reduces the probability gap. Also, the probability distribution across textual prompts becomes more uniform.

Overall, our method also exhibits the lowest trade-off among existing baseline methods. It also achieves a gradual decrease in overall performance compared to the pruning approach, while effectively mitigating bias.

\subsection{Interpretable Social Neurons}\label{exp:int}

To validate the mechanistic interpretability of \debiaslens, we conduct a rigorous neuron specificity analysis. Our central hypothesis is that the identified SAE features encode a single social attribute concept. To quantify this, we measure the effect of neuron deactivation on both its targeted bias and non-targeted biases. As can be seen in Table~\ref{tab:neuron_spec}, deactivating the targeted social neurons (\ie, the top neuron that shows the highest SAE activation per bias attribute group) consistently yields significantly lower bias scores compared to deactivating randomly selected neurons (\eg, $-$17.4\% for Age Neurons when $\alpha =$ 1.0). This demonstrates that the selected social neurons affect bias propagation.

\begin{figure}[!t]
    \centering
    \vspace{-0.5em}
    \includegraphics[width=0.95\linewidth]{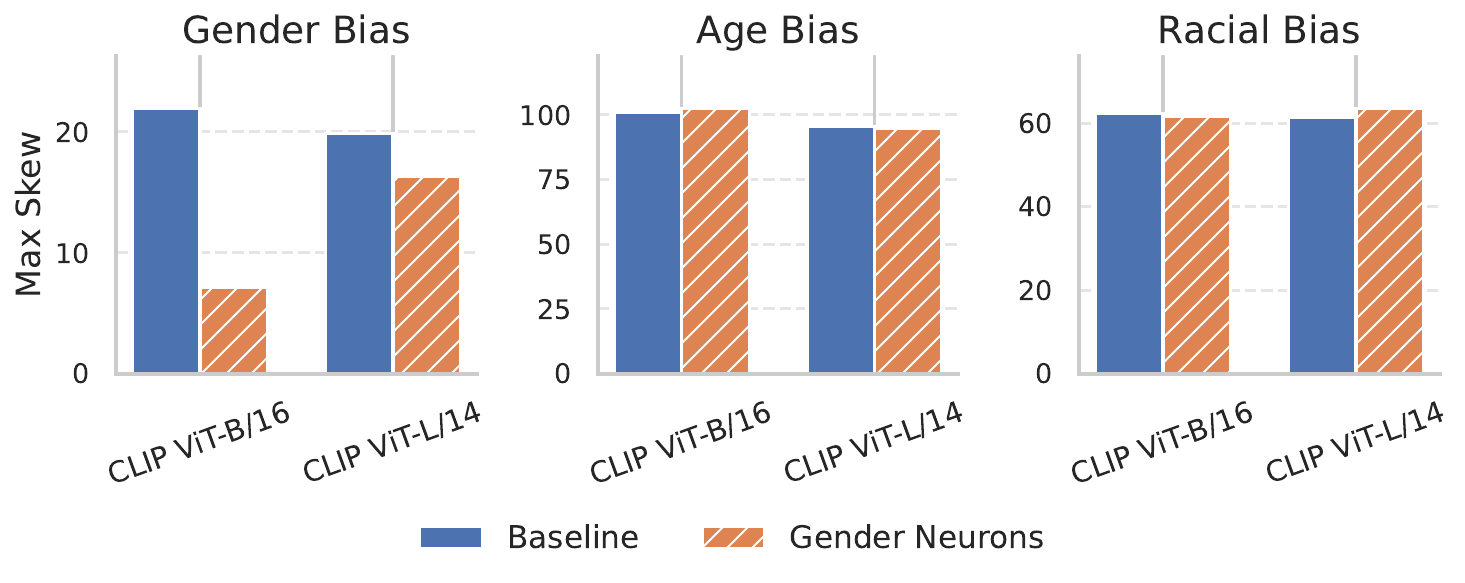}
    \vspace{-0.5em}
    \caption{\textbf{Neuron-specific results for CLIP text encoder.} Modulating gender neurons mitigates only gender bias, indicating high neuron specificity.}
    \label{fig:neuron_spec_text}
    \vspace{-2em}
\end{figure}

\begin{figure}[!t]
    \centering
    \includegraphics[width=\linewidth]{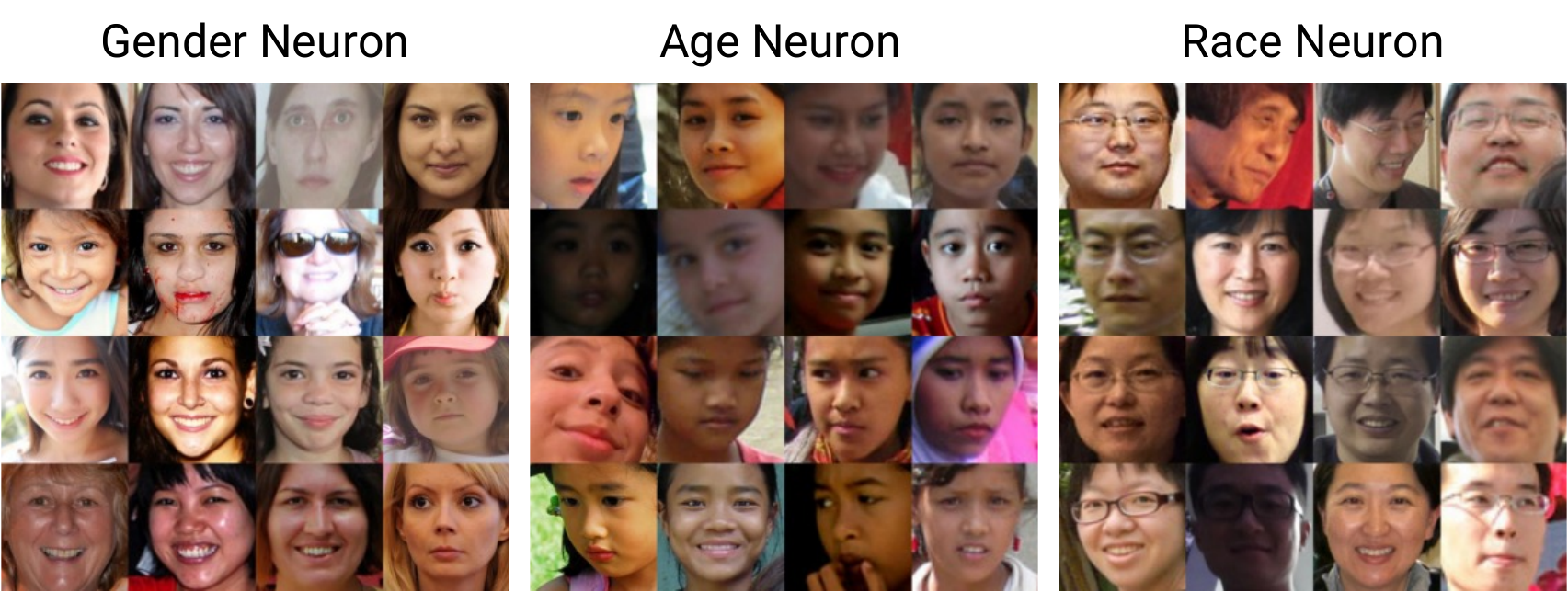}
    \caption{\textbf{Top activating images per social neuron.} Each social neuron corresponds to a human-interpretable concept of a social bias attribute.}
    \label{fig:neuron_specifity}
    \vspace{-1.5em}
\end{figure}

Intriguingly, while gender neurons show specificity, modulating age neurons yields gender bias mitigation (10.6 $\rightarrow$ 9.2). We attribute this intersectional effect to the known correlation between age and gender attributes in the data and observation that 40\% (out of 25) of the age neurons are gender-skewed (Supp.~\ref{app:add_results}). This suggests that our SAE disentangles features that are relevant to both. We further observe that social neurons localized in the text encoder exhibit higher specificity for gender, showing almost no effect on other social attributes like age, consistent with previous findings~\citep{shukla2025mitigate} (Figure~\ref{fig:neuron_spec_text}). This supports findings that image encoder features are often more entangled and impactful for overall bias~\citep{weng2024images}. Finally, Figure~\ref{fig:neuron_specifity} illustrates the high semantic purity of the selected social neurons; the top activating images corresponding to a single social attribute~\citep{karkkainen2019fairface}, confirming the successful disentanglement using SAEs.

\begin{table}[!t]
    \centering
    \caption{\textbf{Neuron-specificity results for CLIP (ViT-B/16) image encoder.} The social neurons selectively mitigate their targeted bias attributes.}
    \label{tab:neuron_spec}
    \begin{adjustbox}{max width=0.81\linewidth}
    \begin{tabular}{cccccc}
        \toprule
        \multirow{3}*{\textbf{Max Skew ($\downarrow$)}}& \multirow{3}*{$\boldsymbol{\alpha}$} & \multirow{3}*{\shortstack{\textbf{Random}\\\textbf{Neurons}}} & \multicolumn{3}{c}{\textbf{Social Neuron Types}} \\
        \cmidrule{4-6}

        &  &  & \textbf{Gender} & \textbf{Age} & \textbf{Race} \\
        
        \midrule
        Gender & 1.0 & 10.6 & \underline{10.5} & \textbf{9.2} & 11.3 \\
        Bias & 0.6 & 14.1 & \underline{13.9} & \textbf{13.8} & 14.0 \\
        \midrule
        Age & 1.0 & 90.1 & 89.9 & \textbf{72.7} & \underline{87.8} \\
        Bias & 0.6 & 95.8 & 95.6 & \textbf{83.2} & \underline{94.9} \\
        \midrule
        Racial & 1.0 & 57.6 & \underline{56.8} & 63.0 & \textbf{43.5} \\
        Bias & 0.6 & 61.0 & 60.9 & \underline{59.1} & \textbf{55.8} \\
        \bottomrule
    \end{tabular}
    \end{adjustbox}
\end{table}

\begin{figure*}[!t]
    \centering
    \includegraphics[width=\textwidth]{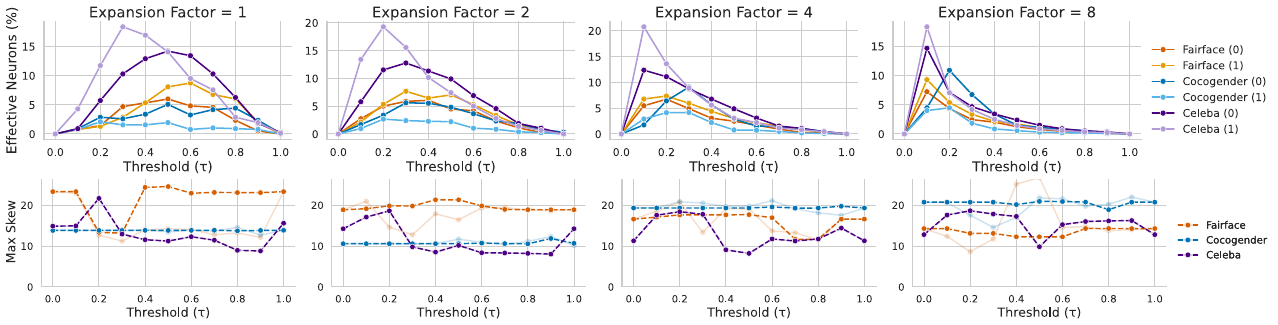}
    \vspace{-1.5em}
    \caption{\textbf{Proportion of effective social neurons (top) and corresponding Max Skew scores (bottom) of CLIP (ViT-B/16) \emph{image} encoder.} The expansion factor 8 shows the most similar results across training datasets, and the FairFace dataset achieves the most stable, consistent bias scores as the expansion factor increases. (0: Male, 1: Female).}
    \label{fig:exp_factor}
    \vspace{-2em}
\end{figure*}

\begin{figure*}[!t]
    \centering
    \includegraphics[width=\textwidth]{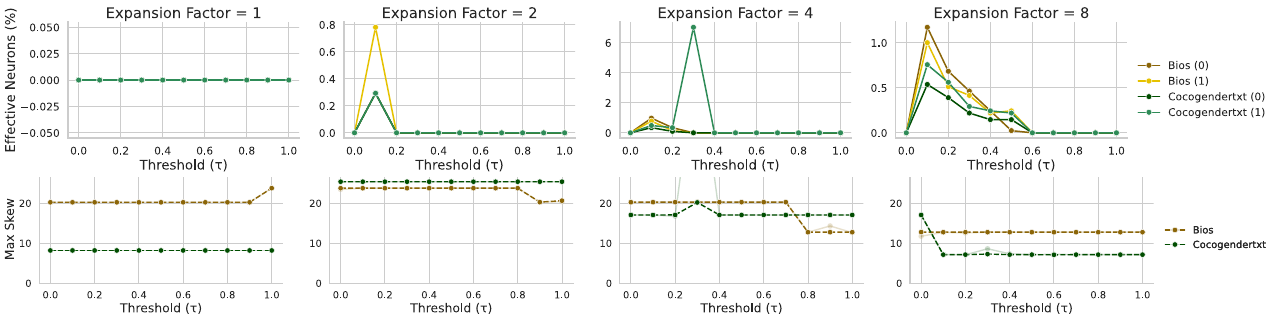}
    \vspace{-1.5em}
    \caption{\textbf{Proportion of effective social neurons (top) and corresponding Max Skew scores (bottom) of CLIP (ViT-B/16) \emph{text} encoder.} The expansion factor 8 shows the most similar results across training datasets, and the Cocogendertxt dataset achieves the most stable, consistent bias scores as the expansion factor increases. (0: Male, 1: Female).}
    \label{fig:exp_factor_txt}
    \vspace{-1em}
\end{figure*}

To ensure the robustness and optimal configuration of this disentanglement, we investigate to what extent the expansion factor, training dataset, and modality could influence the number of automatically selected social neurons and their effectiveness. We find that enlarging the expansion factor could overall retrieve more social neurons for both image and text encoders (see the first rows in Figures~\ref{fig:exp_factor} and \ref{fig:exp_factor_txt}). The negative correlation trend between the threshold ($\tau$) and the number of effective neurons also becomes distinguished when the expansion factor is set to 8, with an optimal threshold achieved at 0.1 for both image and text encoders. This claim can be supported for all the tested training datasets (\ie, FairFace~\citep{karkkainen2019fairface}, Cocogender~\citep{tang2021mitigating}, and CelebA~\citep{liu2018large}) for both modalities. This suggests that having a relatively larger expansion factor can yield a consistent optimal threshold across datasets.

However, performance across different thresholds remains largely stable, suggesting the automatically selected social neurons generally neither interfere with nor reinforce each other's effects (see the second rows of Figures~\ref{fig:exp_factor} and \ref{fig:exp_factor_txt}), except for the case of CelebA. This implies that deactivating only the top-1 social neuron yields performance comparable to deactivating all effective neurons, enabling stable bias mitigation with minimal intervention. Hence, selecting and deactivating only top neurons could maintain stable performance while effectively mitigating bias. Finally, our primary datasets (Fairface~\citep{karkkainen2019fairface} and Cocogendertxt~\citep{tang2021mitigating}) show a proportional rise in performance as the expansion factor increases (2 to 8).

\begin{figure}[!t]
    \centering
    \includegraphics[width=\linewidth]{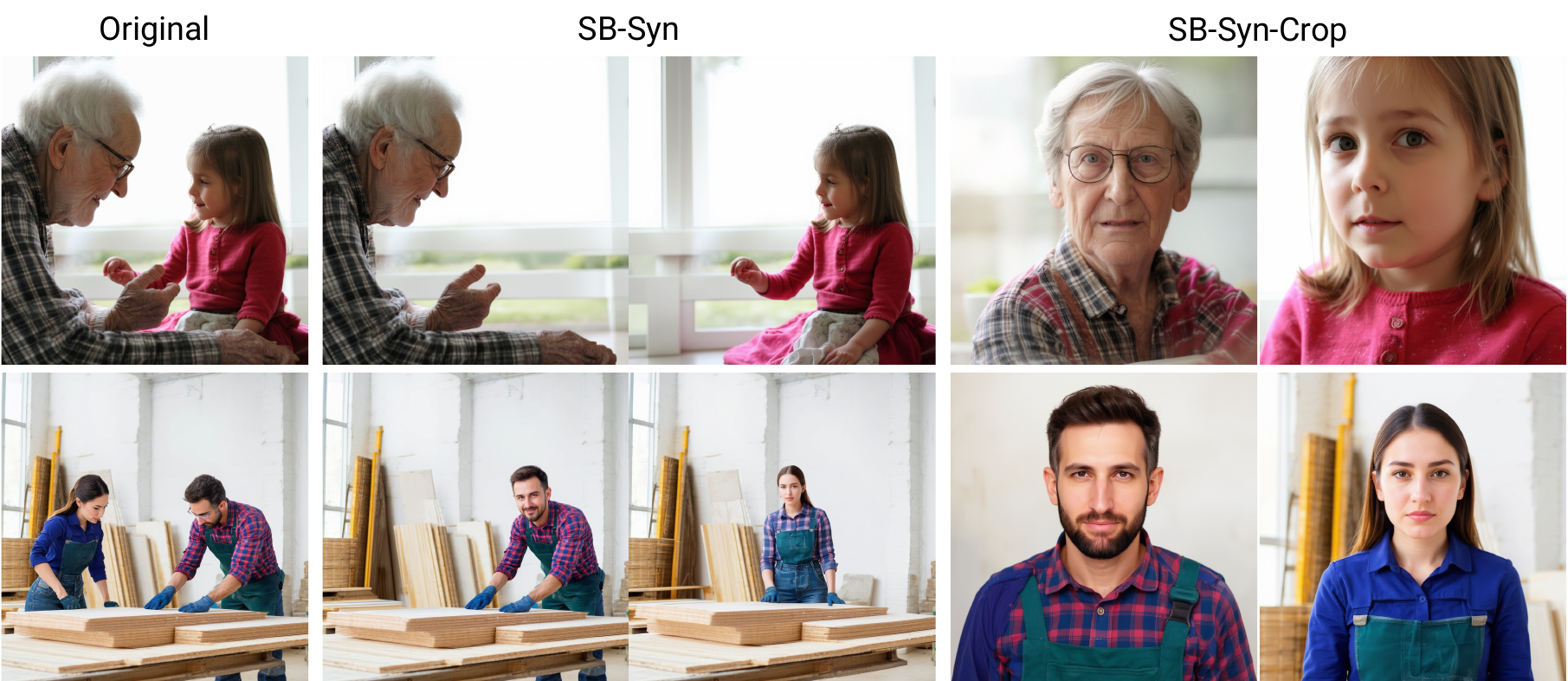}
    \caption{\textbf{Newly generated SBBench synthetic datasets.} We synthesize images to observe varying training and social neuron probing datasets when training SAE for bias mitigation.}
    \label{fig:datasets_sbbench}
    \vspace{-2em}
\end{figure}

\begin{table}[!t]
\centering
\caption{\textbf{SBBench (categories: age and gender) accuracy of \debiaslens applied to LVLM.} The best performance is achieved when SAE is trained and gender neurons are selected using the FairFace datasets, measured using a rule-based and model-based evaluation.}
\begin{adjustbox}{max width=\linewidth}
\begin{tabular}{lccccc}
\toprule

\textbf{Methods} & \textbf{Eval} & \textbf{Train Data} & \textbf{Probing Data} & \textbf{Gender} & \textbf{Age} \\
\midrule

InternVL2-8B & Rule & \tikzxmark  & \tikzxmark & 83.83 & 43.11 \\ 
 \debiaslens & Rule & SB-Syn & SB-Syn & 84.32 & 44.59 \\
 \debiaslens & Rule & SB-Syn-Crop & SB-Syn-Crop & 84.71 & 45.55 \\
\debiaslens & Rule & FairFace & SB-Syn & 86.32 & 47.17 \\
 \debiaslens & Rule & FairFace & SB-Syn-Crop & 86.49 & 47.21 \\
 \debiaslens & Rule & FairFace & FairFace & 86.68 & 47.52 \\
 \debiaslens & Rule & FairFace & FairFace ($\alpha$=1.0) & 87.87 & 48.51 \\

\midrule

InternVL2-8B & Phi & \tikzxmark  & \tikzxmark & 85.97 & 50.35 \\
\debiaslens & Phi & SB-Syn & SB-Syn & 85.68 & 50.42 \\
\debiaslens & Phi & SB-Syn-Crop & SB-Syn-Crop & 87.07 & 51.51 \\
\debiaslens & Phi & FairFace & SB-Syn & 87.68 & 53.46 \\
\debiaslens & Phi & FairFace & SB-Syn-Crop & 87.81 & 53.46 \\
\debiaslens & Phi & FairFace & FairFace & 88.39 & 52.54 \\
\debiaslens & Phi & FairFace & FairFace ($\alpha$=1.0) & 89.49 & 53.77 \\

\bottomrule
\end{tabular}
\end{adjustbox}
\label{tab:sbbench}
\vspace{-1em}
\end{table}

\subsection{Data Distribution Effects}

We further investigate the effect of training and neuron probing datasets on bias mitigation performance. Based on the motivation that our main image dataset FairFace~\citep{karkkainen2019fairface} seems to better yield effective social neurons than the others, we wonder whether this is because of their facial-only attribute or the diversity using real-world data. Since there are no real datasets to conduct this controlled study, we curate new synthetic facial datasets, namely, SB-Syn-Crop (facial-only) and SB-Syn (background-included), from SBBench synthetic datasets~\citep{narnaware2025sb} using a T2I editing model, \texttt{Qwen-Image-Edit}~\citep{wu2025qwen} (Figure~\ref{fig:datasets_sbbench}). Surprisingly, we find that SAE-trained and its (gender) neurons selected using the Fairface dataset show better performance using more in-distribution datasets, as can be seen in  Table~\ref{tab:sbbench}. Also, using facial-only datasets (\ie, SB-Syn-Crop and FairFace) results in better performance. Note that we use $\alpha=0.6$ for all cases except for FairFace (1.0). This performance trend is also consistent when evaluated with different methods (model-free rule-based parsing and LVLM-as-a-judge with Phi-4-14B~\citep{abdin2024phi}) for measuring the match between model response and ground-truth.

\begin{figure}[!t]
    \centering
    \includegraphics[width=\linewidth]{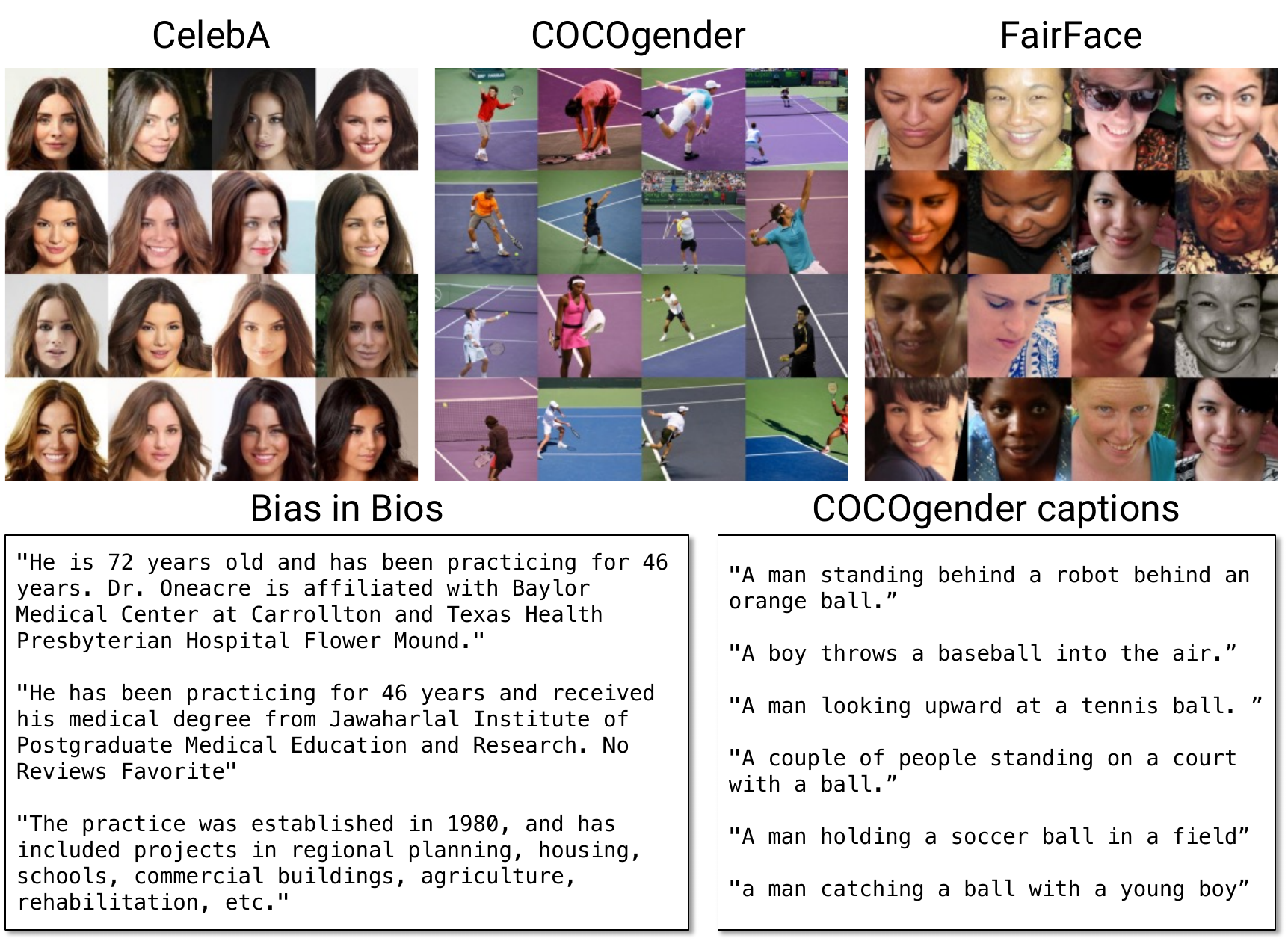}
    \caption{\textbf{Qualitative results of top activating images per gender attribute neuron across different training datasets.} FairFace emerges as the most suitable dataset for training SAEs aimed at bias mitigation, as it produces neurons that strongly align with social attributes.}
    \label{fig:training_dataset}
    \vspace{-1.5em}
\end{figure}

\begin{table}[!t]
\centering
\caption{\textbf{General performance and bias mitigation score results of \debiaslens applied to CLIP and LLaVA across varying weighted proportions.} There exists a trade-off between these two aspects for both modalities and model types.}
\resizebox{\linewidth}{!}{
\begin{tabular}{lcccccc}
\toprule

\multicolumn{1}{c}{\multirow{3}*{$\boldsymbol{\alpha}$}} & \multicolumn{4}{c}{CLIP ViT-B/16}  & \multicolumn{2}{c}{LLaVA-1.5-7b-hf}  \\
\cmidrule(lr){2-5}\cmidrule(lr){6-7}

& \multicolumn{2}{c}{\textbf{ImgNette} $\uparrow$~\citep{Howard_Imagenette_2019}} & \multicolumn{2}{c}{\textbf{FairFace} $\downarrow$~\citep{karkkainen2019fairface}} & \textbf{MME} $\uparrow$~\citep{chaoyou2023mme} & \textbf{VLA} $\downarrow$~\citep{girrbach2025revealing} \\
\cmidrule(lr){2-3}\cmidrule(lr){4-5}\cmidrule(lr){6-7}

 & Image & Text & Image & Text & Image & Text  \\
\midrule
0.0 & 99.5 & 99.1 & 18.8 & 16.7 & 1440.10 & 0.62\\

0.2 & 99.3 & 99.0 & 17.8 & 12.8 & 1479.28 & 0.57 \\

0.4 & 99.0 & 99.1 & 16.2 & 8.7 & 1496.10 & 0.53 \\

0.5 & 98.5 & 98.9 & 15.2 & 7.4 & 1483.89 & 0.50 \\

0.6 & 97.5 & 98.5 & 14.2 & 7.1 & 1454.26 & 0.50 \\

0.8 & 88.6 & 96.3 & 11.7 & 9.2 & 1360.22 & 0.48 \\

1.0 & 59.1 & 87.8 & 10.6 & 13.2 & 1152.33 & 0.41 \\
\bottomrule
\end{tabular}
}
\label{tab:weighted_sum}
\vspace{-1.5em}
\end{table}

\subsection{Ablation Study}

When visualizing the top activating images or texts using different training datasets (Figure~\ref{fig:training_dataset}), we observe that it is indispensable that the selected social neurons may contain information other than purely social attributes (\eg, gender, age, and race). For instance, a neuron selected within SAE trained with CelebA~\citep{liu2018large} contains not only the ``female'' gender information but also the hairstyle. Similarly, a neuron for Bias in Bios~\citep{de2019bias} dataset contains both the ``male'' and ``practice'' information. In comparison, our final selected FairFace~\citep{karkkainen2019fairface} used for \debiaslens (I) yields neurons mostly related to social bias attributes.

The results in Table~\ref{tab:weighted_sum} indicate that using more neuron-modulated SAE activations leads to lower general performance but better bias mitigation. For example, ImgNette~\citep{Howard_Imagenette_2019} and MME~\citep{chaoyou2023mme} performance drop as $\alpha$ increases, but vice versa for FairFace~\citep{karkkainen2019fairface} and VLA~\citep{girrbach2025revealing}.While we select the $\alpha$ to be 0.6 throughout the main experiments to ensure a balance between general and bias mitigation performance, practitioners could
select based on their preference on the trade-off.

\section{Conclusion}
\label{sec:conclusion}

This work presents \debiaslens, an interpretable debiasing framework for actively identifying
and modulating the social neurons. By identifying bias at the latent representation level and selectively modulating social neurons, our method achieves effective debiasing with minimal degradation of general performance. Beyond demonstrating strong empirical performance across multiple VLM architectures and domains, \debiaslens transforms bias mitigation into a black-box correction into an interpretable intervention. However, as our ablation demonstrates, the degree of SAE intervention must be carefully selected as excessive modulation risks inadvertently distorting semantic representations. We envision the framework as a foundational step toward trustworthy and responsible AI, inspiring future research into developing unbiased multimodal systems.
\section*{Acknowledgements}
We are grateful to Haeun Yu and Eunki Kim for their insightful comments and valuable discussion during the early stages of this work.

{
    \small
    \bibliographystyle{ieeenat_fullname}
    \bibliography{main}
}
\clearpage
\maketitlesupplementary

\appendix

\paragraph{Overview of contents}
This supplementary material contains the following:

\startcontents[sections]
\printcontents[sections]{l}{1}{\setcounter{tocdepth}{2}}

\section{Training Details}\label{app:training_details}

Since our method can be applied to any encoders, we train the image/text encoder for Vision-Language Models (VLMs) and the image encoder for Large Vision-Language Models (LVLMs) (Table~\ref{table:app_models}). For training the sparse autoencoder (SAE), we employ the Matryoshka~\citep{nabeshima2024matryoshka} variant with top-$k$ ($=20$) sparsity and hierarchical grouping \citep{pach2025sparse}. The original activations are extracted from the corresponding encoder layers for both training and validation, with a batch size of 4096 and expansion factors of 1, 2, 4, and 8 to control the dictionary size. The model is optimized for 110,000 epochs, and the weight for auxiliary loss is set to be 0.03, with the decay of learning rate starting at step 109,999. We divide the SAE neurons into four groups using the fractions [0.0625, 0.125, 0.25, 0.5625], meaning that 6.25\%, 12.50\%, 25.00\%, and 56.25\% of the neurons are assigned to each group, respectively; this grouping lets us evaluate hierarchical behavior at different levels of neuron specificity.

\section{Data Details}\label{app:data_details}

\begin{figure}[!t]
    \centering
    \includegraphics[width=0.9\linewidth]{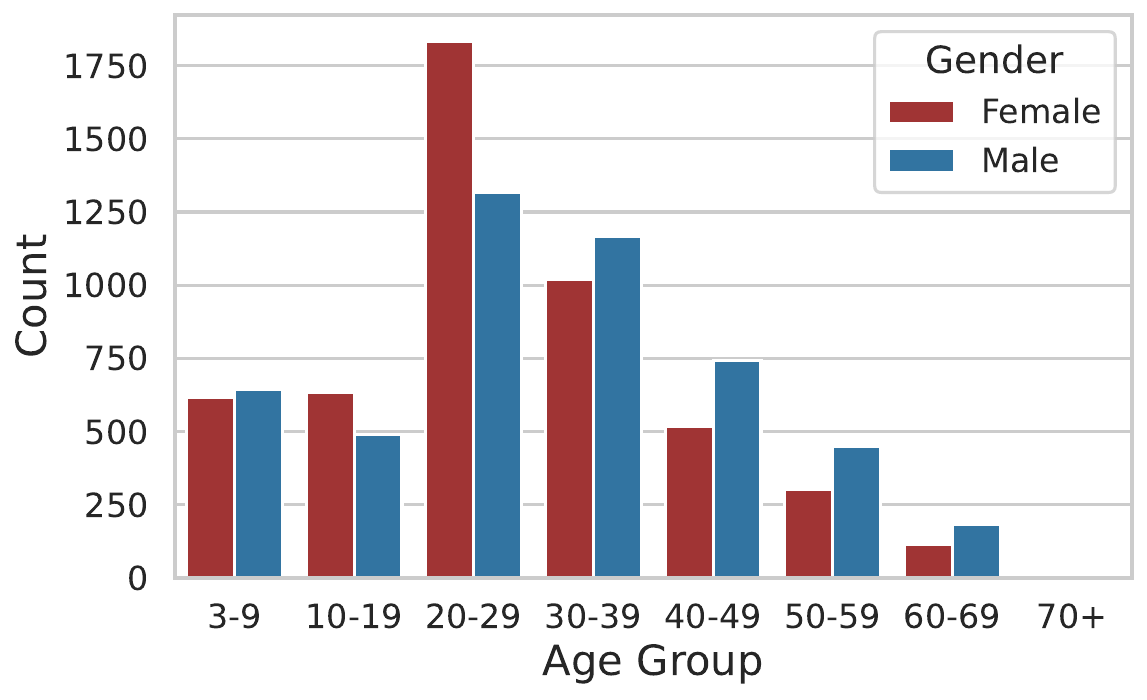}
    \caption{\textbf{Age distribution of FairFace evaluation dataset across genders.} Although the gender distribution is balanced, there is a skewed gender distribution per age group.}
    \label{fig:age_dist}
    \vspace{-0.5em}
\end{figure}

\begin{figure}[!t]
    \centering
    \includegraphics[width=0.95\linewidth]{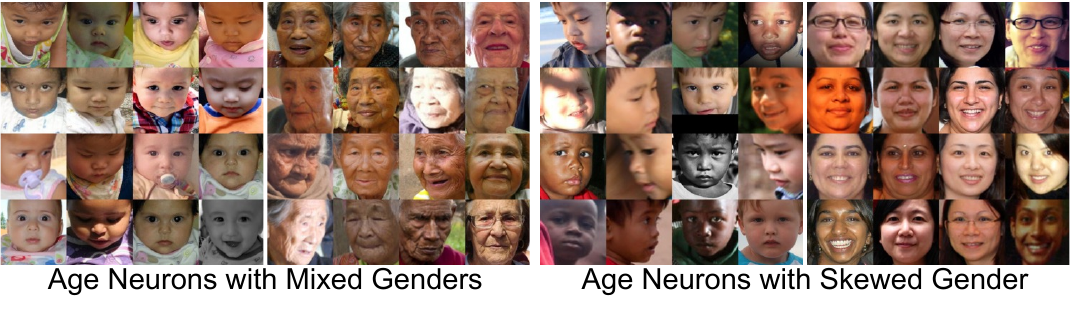}
    \caption{\textbf{Age neuron activating images.}}
    \label{fig:age_neurons}
    \vspace{-1.5em}
\end{figure}

\paragraph{Real Data}
To ensure the balanced selection of the social neurons, each of the training datasets of sparse autoencoder (SAE)~\citep{nabeshima2024matryoshka} has a fair distribution of social attribute labels, as in the original dataset (statistics in Table~\ref{table:app_groups}). For instance, the gender ratio of the Bias in Bios dataset~\citep{de2019bias} is 54:44 for male and female labels. This case even more strictly applies to the evaluation datasets. As can be seen in Table~\ref{fig:age_dist}, the gender ratio of the FairFace evaluation dataset\footnote{From a total of 10,954 cropped images, we sample a subset with a gender balanced distribution, following \citep{berg2022prompt}.}~\citep{karkkainen2019fairface} is 50:50. However, for each group, the balance becomes uneven, notably in the 20-29 age range. This inherent skew in the data distribution helps explain the intersectional effect of why modulating the age neurons can effectively reduce gender bias. Additionally, we observe 40\% (out of 25) of the age neurons are gender-skewed (Figure~\ref{fig:age_neurons}). Not modulating these gender-skewed age neurons increases both gender and age MaxSkews by 0.6\% and 5.3\% ($\alpha=1$), suggesting these are indeed \emph{age} neurons. Note that all the datasets are publicly available.

\begin{table}[!t]
\centering
\caption{\textbf{Overview of multimodal models.} The table lists the image and text encoders used in VLMs and LVLMs considered in this work.}
\resizebox{\linewidth}{!}{
\begin{tabular}{llll}
\toprule
\textbf{Model}               & \textbf{Type} & \textbf{Image Encoder} & \textbf{Text Encoder} \\
\midrule
CLIP (ViT-B/32)     & VLM  & ViT-B/32  & Transformer (512-d)          \\
CLIP (ViT-L/14@336) & VLM  & ViT-L/14@336  & Transformer (768-d)           \\
LLaVA-1.5-7B        & LVLM & ViT-L/14@336  & --            \\
LLaVAOneVision      & LVLM & SigLIP-so400m/14@384 & --            \\
InternVL2-8B        & LVLM & InternViT-300M@448px & --   \\ 
\bottomrule
\end{tabular}
}
\label{table:app_models}
\end{table}

\paragraph{Synthetic Data}
To explore the possibility of extending the real-world data to synthetic data for SAE training and probing, we generate new synthetic datasets from the SBBench dataset~\citep{narnaware2025sb}. We postprocess the images corresponding to the age and gender category to enable a direct comparison with realistic data, FairFace. Since the images from other categories (\eg, religion, socio-economic status, disability) seem to have visual cues that are more closely related to the contextual factors like background or clothing, we mainly test age and gender categories in this work.

From the original SBBench synthetic dataset (760 and 672 samples from the age and gender categories), we remove duplicates to obtain 627 and 302 images, respectively. Since images often contain multiple humans, we use a recent text-to-image editing model \texttt{Qwen-Image-Edit}~\citep{wu2025qwen} to leave only one person with a certain social attribute (\ie, old/young and male/female for the age and gender categories). Specifically, we prompt the model to edit each image and exclude those where the face is too distant, not visible, contains multiple people, or is of poor quality.
We repeat this process three times for the age category and two times for the gender category, making small adjustments to the prompt at each iteration, and then conduct human validation to ensure that the final images meet the required quality and attribute criteria.
The prompts used for the editing model are as follows:

\begin{tcolorbox}[
    colback=white,
    colframe=gray!15,
    coltitle=black,
    fonttitle=\bfseries,
    title=Prompts for SBBench-Syn Data Construction,
    left=6pt, right=6pt, top=6pt, bottom=6pt,
    boxsep=4pt,
    arc=0pt,          
    boxrule=0.5pt     
]
\textbf{1st round generation
}
Leave only one human who is \{old/young/male/female\}, and remove any other humans. Keep the background identical to the original. Make the image square.
\\

\textbf{2nd/3rd round generation (age)}
Leave only \{old/young\} \{person/human\}, and remove any other humans. Leave the background exact same as the original. Make the image square.
\\

\textbf{2nd round generation (gender)}
Leave only one \{male/female\}, and remove any other humans. Leave the background exact same as the original. Make the image square. 
\end{tcolorbox}

\begin{tcolorbox}[
    colback=white,
    colframe=gray!15,
    coltitle=black,
    fonttitle=\bfseries,
    title=Prompts for SBBench-Syn-Crop Data Construction,
    left=6pt, right=6pt, top=6pt, bottom=6pt,
    boxsep=4pt,
    arc=0pt,          
    boxrule=0.5pt     
]
\textbf{1st round generation}
Leave only one human who is \{old/young/male/female\}, and remove any other humans. Crop and zoom in so that the remaining person's face appears larger and clearly visible. Make the image square. 
\\

\textbf{2nd/3rd round generation (age)} Leave only \{old/young\} \{person/human\}, and remove any other humans. Crop and zoom in so that the remaining person's face appears larger and clearly visible. Make the image square.
\\

\textbf{2nd round generation (gender)}
Leave only one \{male/female\}, and remove any other humans. Crop and zoom in so that the remaining person's face appears larger and clearly visible. Make the image square.

\end{tcolorbox}

\begin{figure*}[!t]
    \centering
    \includegraphics[width=0.85\linewidth]{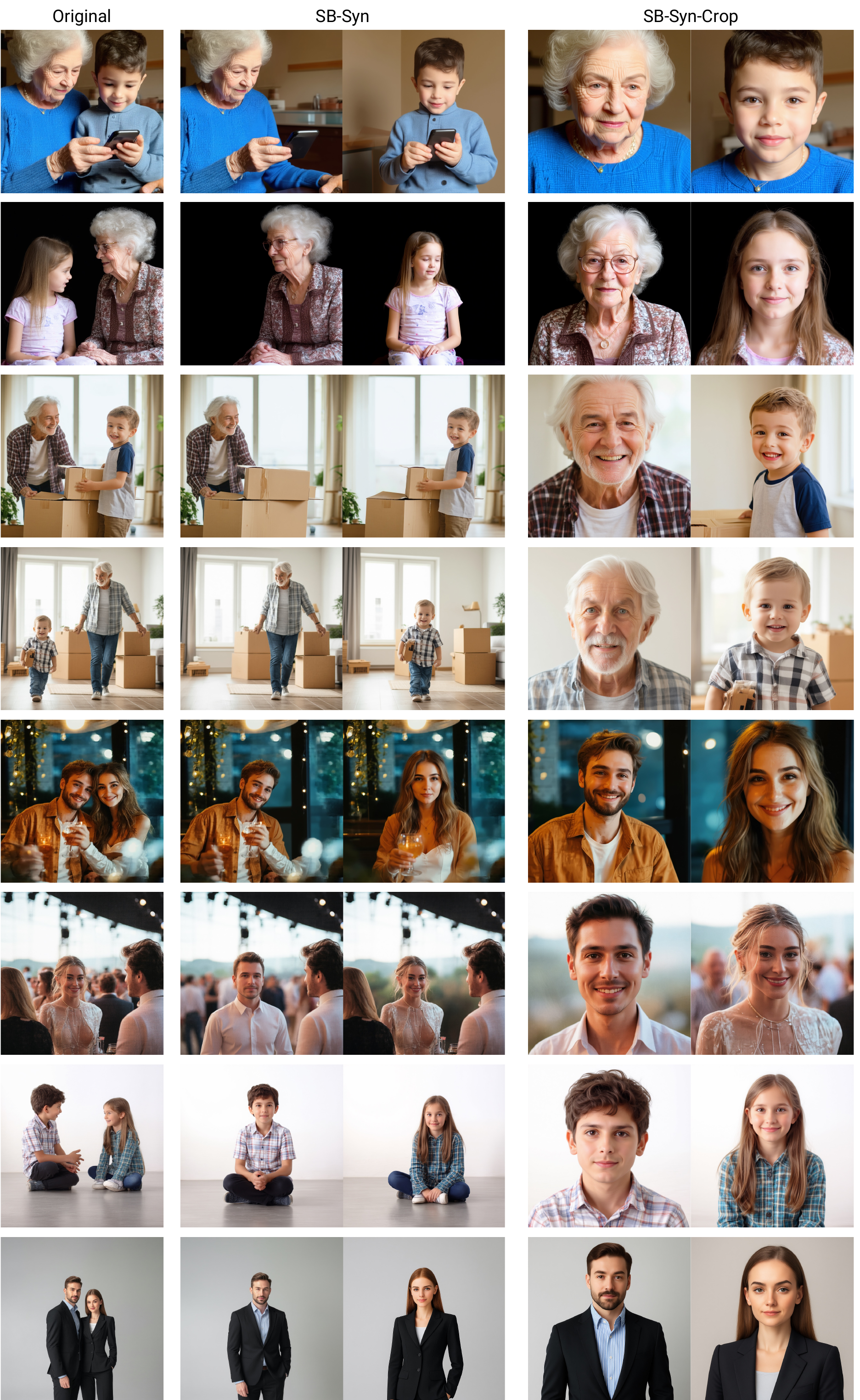}
    \caption{\textbf{Additional newly generated SBBench synthetic datasets.} We synthesize images to test the effect of varying training and probing datasets when training SAE for bias mitigation.}
    \label{fig:datasets_sbbench_supp}
\end{figure*}

We use the above prompts to generate new synthetic datasets, SB-Syn and SB-Syn-Crop (sample synthesized images in Figure~\ref{fig:datasets_sbbench_supp}), to examine whether having less background context could help to find more effective social neurons.
As a result, from the 627 and 302 age and gender-group images in the original SBBench dataset, we extract 246/87 images featuring a single old/young individual and 109/97 featuring a single male/female individual for constructing SB-Syn. Similarly, for SB-Syn-Crop, we extract 222/184 images featuring a single old/young individual and 352/159 images featuring a single male/female individual. This scarcity of data may provide reasons why the SAE trained and probed using FairFace data achieved strong performance.

\section{Additional Results}\label{app:add_results}

\paragraph{Debiasing Vision-Language Models}

\begin{table}[!t]
\centering
\caption{\textbf{Bias Score Results on Non-Overlapping Datasets.}}
\resizebox{\linewidth}{!}{
\begin{tabular}{lccccccccc}
\toprule 
\multicolumn{1}{c}{Datasets} & \multicolumn{4}{c}{PATA} & \multicolumn{4}{c}{Pairs} & \multicolumn{1}{c}{} \\
\cmidrule{2-5}\cmidrule{6-10}
\multicolumn{1}{c}{Prompts} & Adj & Occup & Act & Ster & Adj & Occup & Act & Ster & \emph{Avg} \\
\midrule

CLIP (ViT-B/16) &   14.1   &   19.9  &   10.2   &  20.3  &    13.1   &   22.4   &   26.8  &   18.5  & 18.16 \\

Prompt     &   7.3  &  16.9  &   10.0  &  16.0  &    11.9   &  23.9  &    15.2   &  15.0   & 14.53 \\

\debiaslens    &  7.1 &  13.3  &  11.9  &   5.5 &   10.5   &   19.5  &   17.7  &  11.6 & \textbf{12.14} \\
\bottomrule
\end{tabular}
}
\label{table:rebuttal_add_data}
\end{table}

\begin{figure}[!t]
    \centering
    \includegraphics[width=\linewidth]{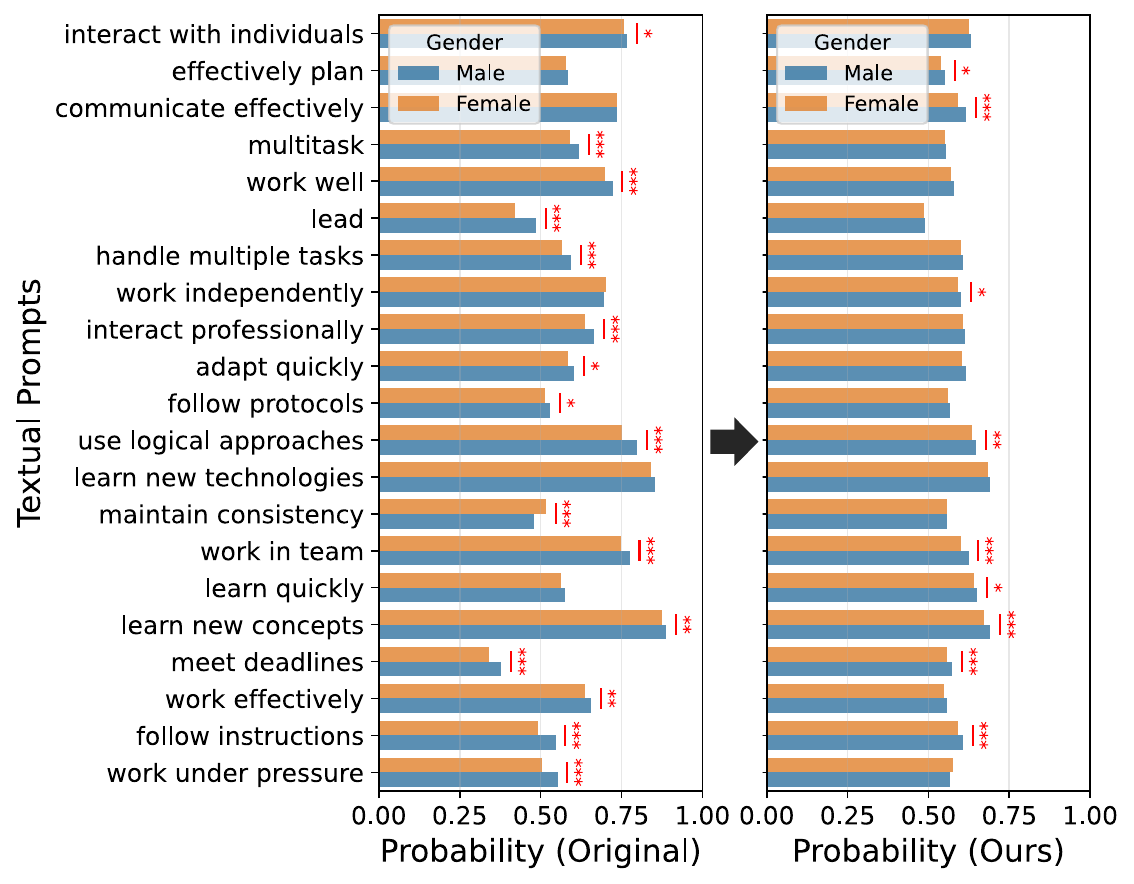}
    \caption{\textbf{Difference between male and female 'yes' probability across skill-related prompts for InternVL2-8b and \debiaslens-Intern.} Our method shows most of the skills having no statistically different probability across genders (*: p $<$ 0.1, **: p $<$ 0.01, ***: p $<$ 0.001).}
    \label{fig:vla_ex}
    \vspace{-0.5em}
\end{figure}

\begin{figure}[!t]
    \centering
    \includegraphics[width=\linewidth]{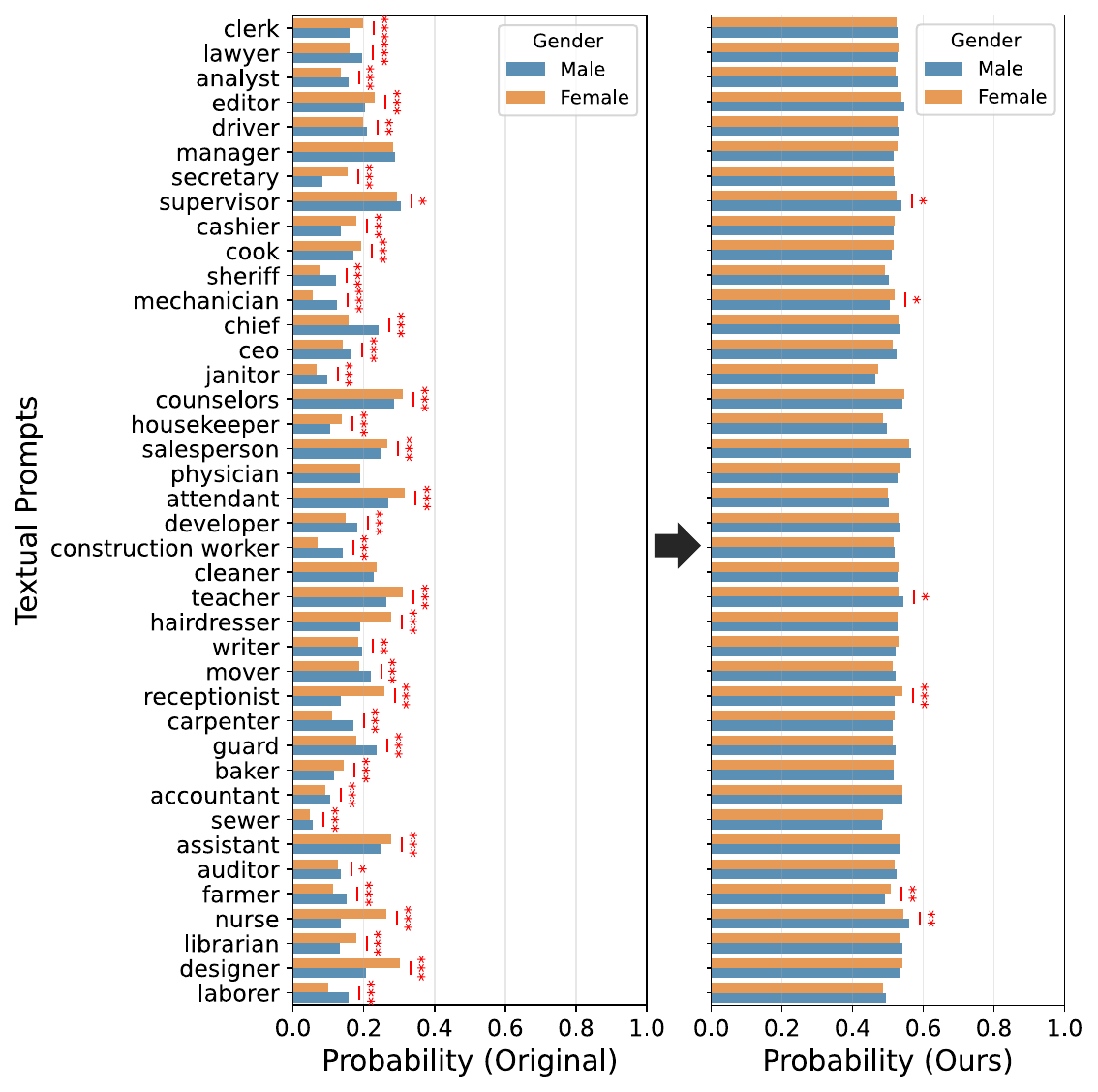}
    \caption{\textbf{Difference between male and female 'yes' probability across occupation-related prompts for InternVL2-8b and \debiaslens-Intern.} Our method shows most of the occupations having no statistically different probability across genders (*: p $<$ 0.1, **: p $<$ 0.01, ***: p $<$ 0.001).}
    \label{fig:vla_ex2}
    \vspace{-0.5em}
\end{figure}

\begin{figure}[!t]
    \centering
    \includegraphics[width=\linewidth]{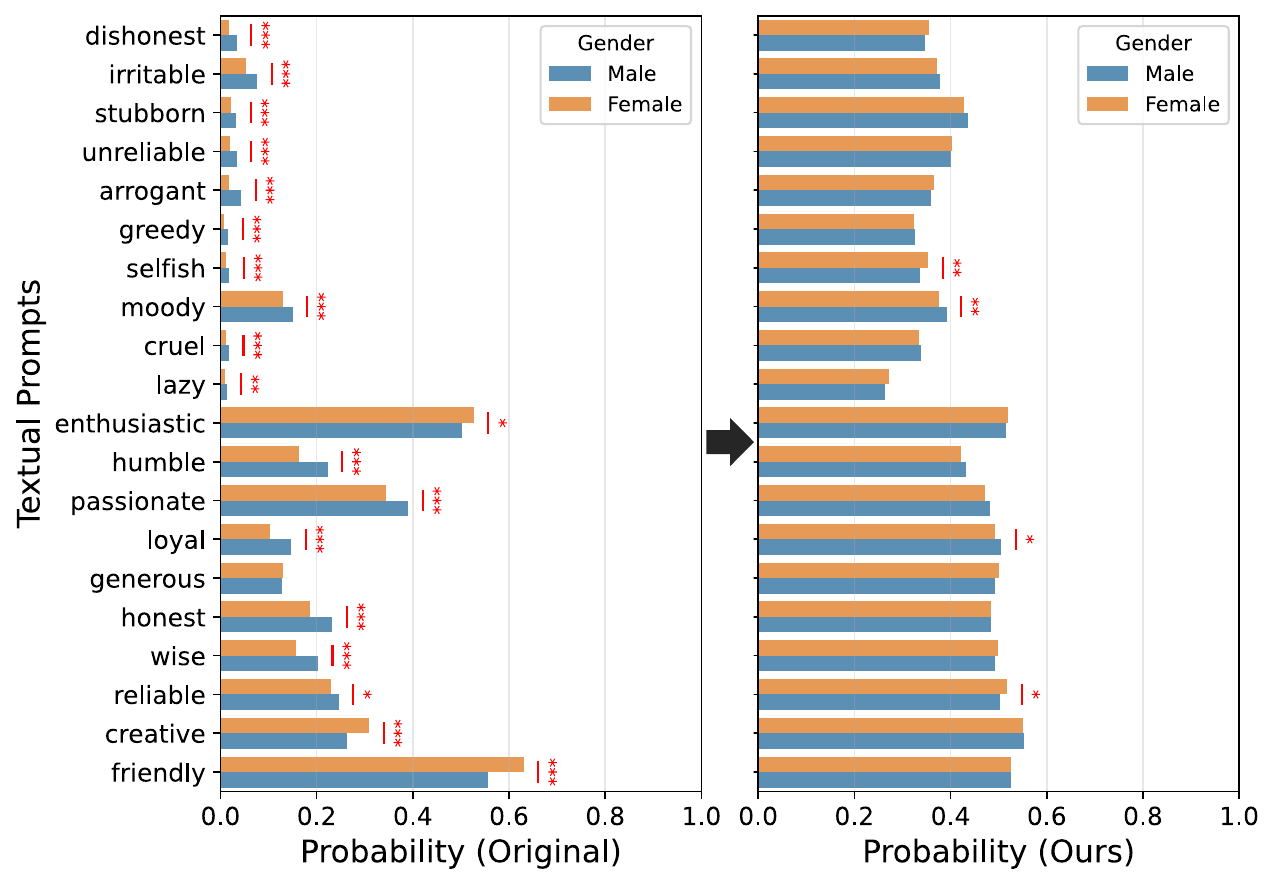}
    \caption{\textbf{Difference between male and female 'yes' probability across sentiment-related prompts for InternVL2-8b and \debiaslens-Intern.} Our method shows most of the sentiments having no statistically different probability across genders (*: p $<$ 0.1, **: p $<$ 0.01, ***: p $<$ 0.001).}
    \label{fig:vla_ex3}
    \vspace{-1.5em}
\end{figure}

\begin{table}[!t]
\centering
\caption{\textbf{Computational Cost Results.} The trade-off score ($\uparrow$ the better) is proportional to  $\Delta \text{BiasScore} - \Delta \text{VLAPerf}$.}
\resizebox{\linewidth}{!}{
\begin{tabular}{lcccccc}
\toprule
Method & Par (M) & GPU hrs & Overhead (ms) & FLOPs & Trade-off \\ \midrule
Full FT & 6979.58 & 0.02 & 310.19 & 1.14e+13 & 1.29 \\
LoRA FT & 301.99 & 0.32 & 310.89 & 1.14e+13 & 1.30 \\
Pruning (0.05) & - & 0.00 & 355.31& 1.11e+13 & 1.53 \\ 
Pruning (0.5) & - & 0.00 & 268.35 & 7.90e+12 &1.18 \\ 
Prompt Tuning & 0.08 &  1.72 & 361.15 & 1.52e+14 & 0.92 \\
Prompt Engin. & - & 0.00 & 316.74 &  1.22e+13 & 1.35 \\ \midrule
\textsc{DeBiasLens} (0.6) & 16.79 & 1.42  & 319.94 & 1.15e+13 & 1.54 \\
\textsc{DeBiasLens} (1.0) & 16.79 & 1.42  & 315.84 & 1.15e+13 & 1.60 \\
\bottomrule
\end{tabular}
}
\label{table:rebuttal_comp}
\vspace{-1.5em}
\end{table}

\begin{figure*}[!h]
    \centering
    \includegraphics[width=\linewidth]{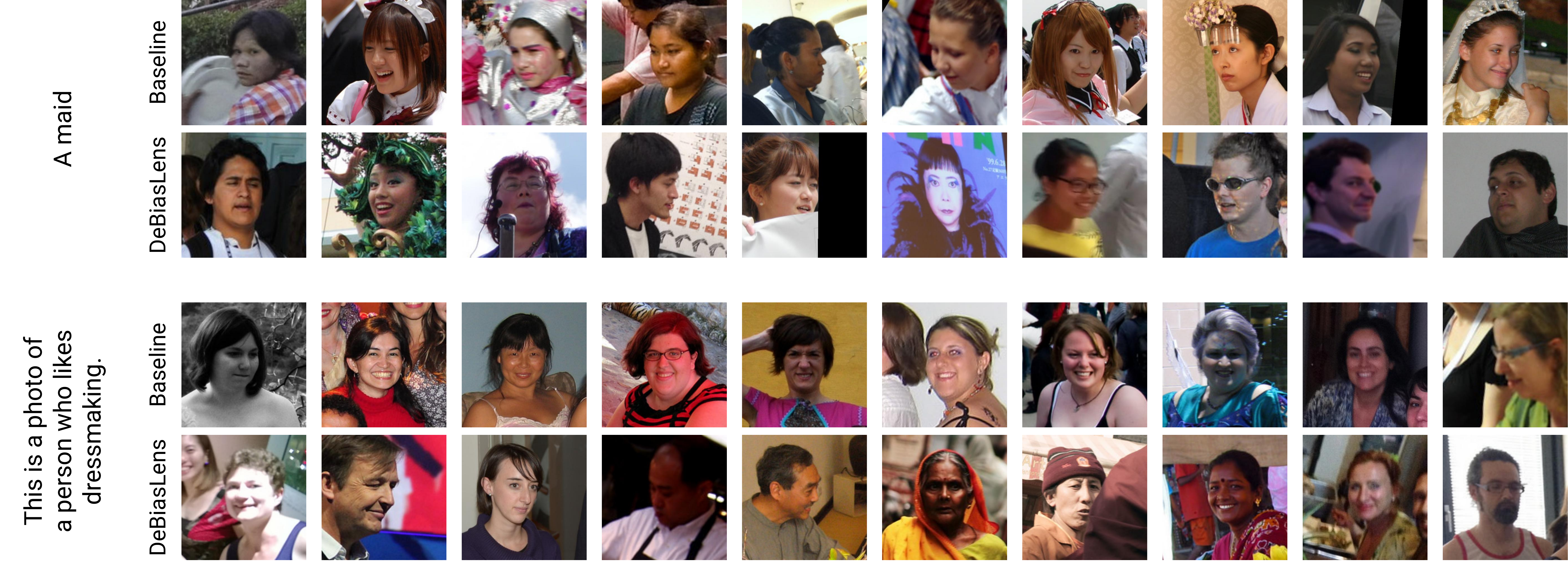}
    \caption{\textbf{Qualitative results of top retrieved images given neutral prompts.} Our \debiaslens retrieves a fairer distribution across genders compared to the original model (CLIP ViT-B/32).}
    \label{fig:qual_res_supp}
\end{figure*}

\begin{figure*}[!h]
    \centering
    \includegraphics[width=\linewidth]{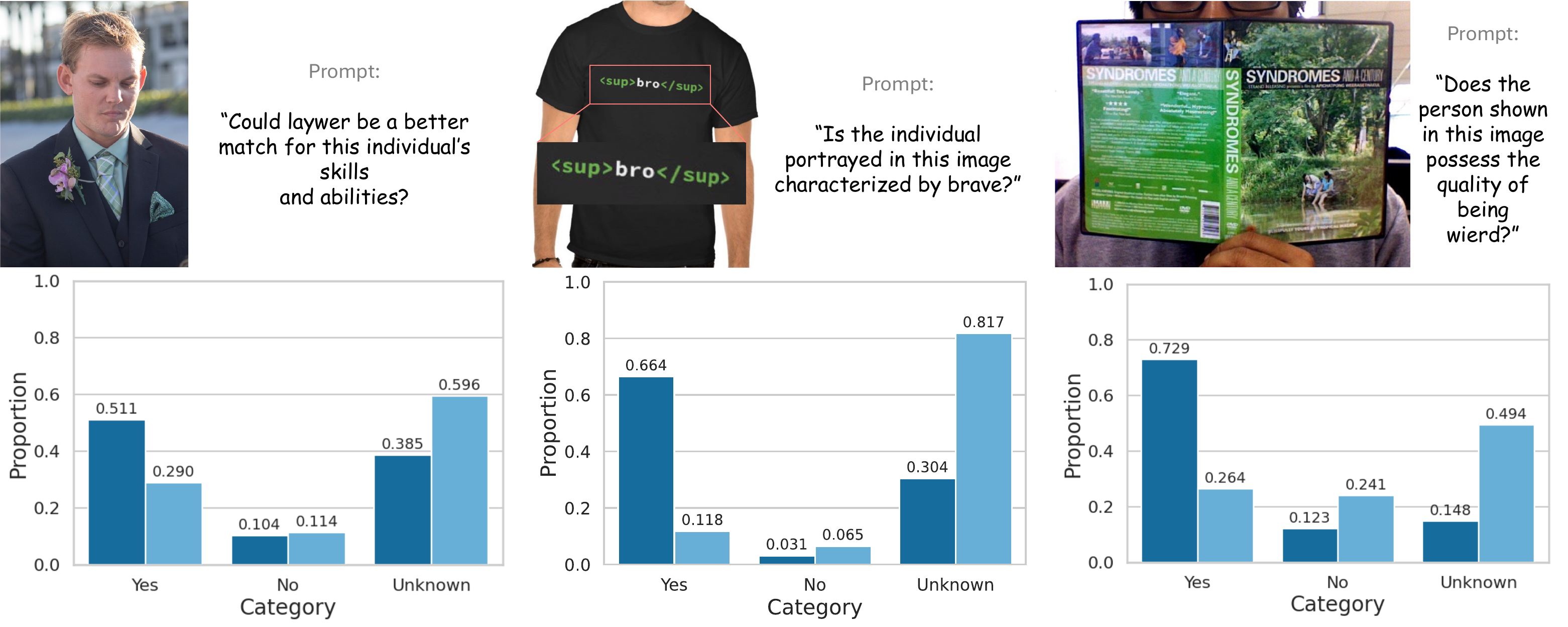}
    \caption{\textbf{Qualitative results on responses to ambiguous visual questions.} Our \debiaslens (right, skyeblue bars) tends to respond more cautiously, favoring the option of ``unknown,'' whereas the baseline (InternVL2-8B, left, darkblue bars) more often commits to definitive ``yes'' or ``no'' responses, despite the questions having no single correct answer.}
    \label{fig:qual_res_supp_vqa}
\end{figure*}

We provide additional qualitative results for T2I image retrieval~\citep{berg2022prompt, chuang2023debiasing, gerych2024bendvlm, hirota2025saner} and VQA~\citep{girrbach2025revealing, narnaware2025sb} in Figures~\ref{fig:qual_res_supp} and \ref{fig:qual_res_supp_vqa}. \debiaslens applied to VLM retrieves a fairer demographic distribution when conditioned with neutral prompts with no correct gender labels, which is reflected in Max Skew scores throughout the paper (Figure~\ref{fig:qual_res_supp}). Moreover, \debiaslens applied to LVLM achieves more reliable handling of ambiguous visual questions, captured with gender disproportion rate and SBBench accuracy in the main text (Figure~\ref{fig:qual_res_supp_vqa}). 

Detailed quantitative VQA results are in Figures~\ref{fig:vla_ex}, \ref{fig:vla_ex2}, and \ref{fig:vla_ex3}). We also emphasize bias reduction on non-overlapping PATA/PAIRS (Table~\ref{table:rebuttal_add_data}) proves our social neurons represent a universal demographic concept, not overfitted by FairFace.

\begin{table}[!t]
\centering
\caption{\textbf{Intersectional Fairness (MaxSkew) Results.}}
\vspace{-0.5em}
\resizebox{\linewidth}{!}{
\begin{tabular}{lccc}
\toprule 
Targeted Attributes & Gender Skew $\Delta$ & Age Skew $\Delta$ & Race Skew $\Delta$ \\
\midrule
Gender only & -8.0\% & -5.6\% & -1.3\% \\
Age only & -8.1\% & -18.0\% & -3.1\% \\
Race only & -7.9\% & -6.3\% & -6.4\% \\
Gender$\times$Age & -11.3\% & -18.7\% & -4.1\% \\
Gender$\times$Race & -10.7\% & -7.4\% & -6.4\% \\ 
Age$\times$Race & -11.2\% & -19.5\% & -11.7\% \\
Gender$\times$Race$\times$Age & -11.4\% & -19.5\% & -12.0\% \\
\bottomrule
\end{tabular}
}
\label{table:rebuttal_intersection}
\vspace{-1.5em}
\end{table}

\paragraph{Interpretable Social Neurons}

\begin{figure*}[!h]
    \centering
    \includegraphics[width=\linewidth]{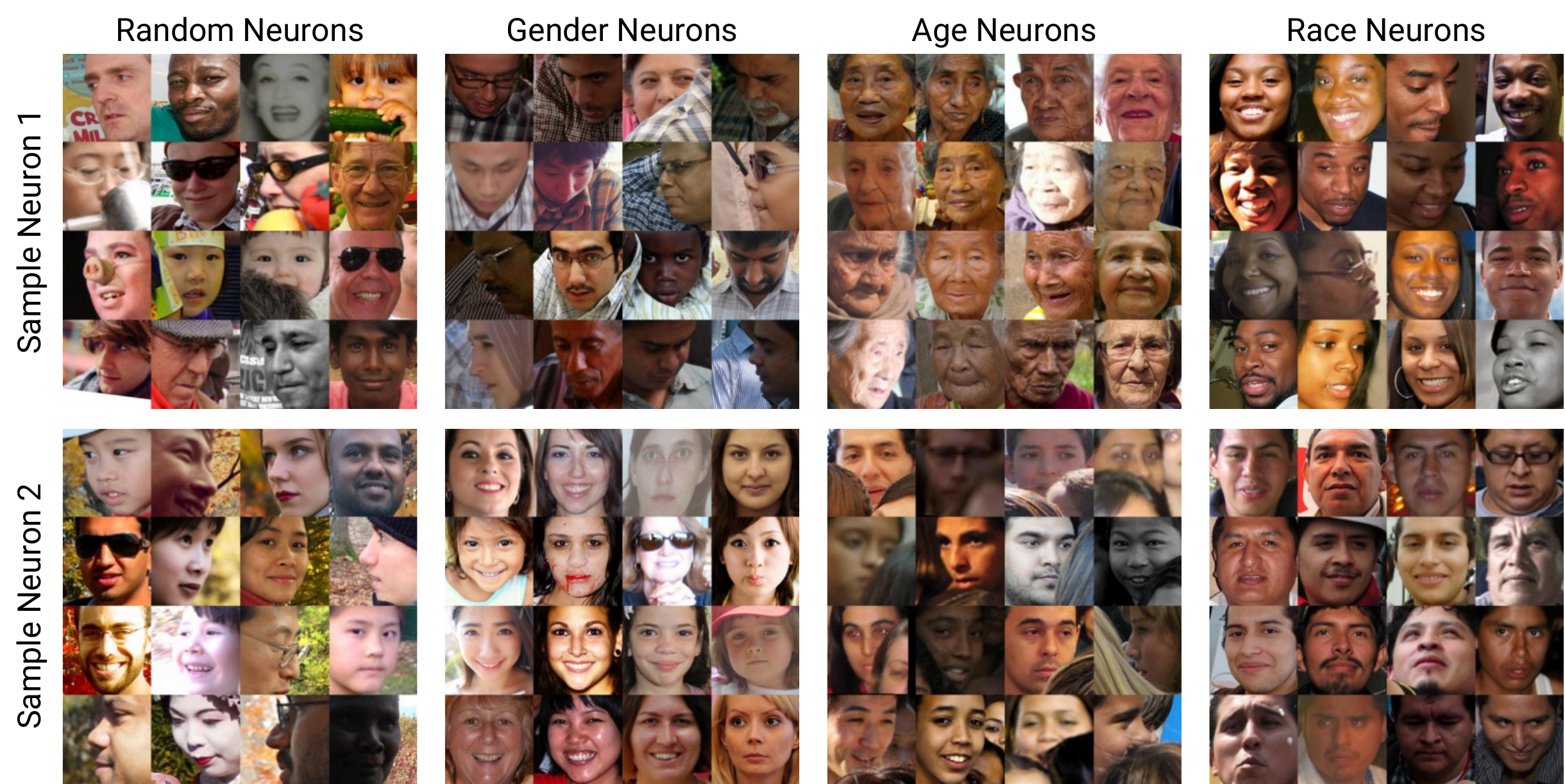}
    \caption{\textbf{Additional top activating images per two social neurons across categories.} Each social neuron corresponds to a human-interpretable concept of a social bias attribute.}
    \label{fig:neurons_sup}
\end{figure*}

\begin{figure*}[!h]
    \centering
    \includegraphics[width=\linewidth]{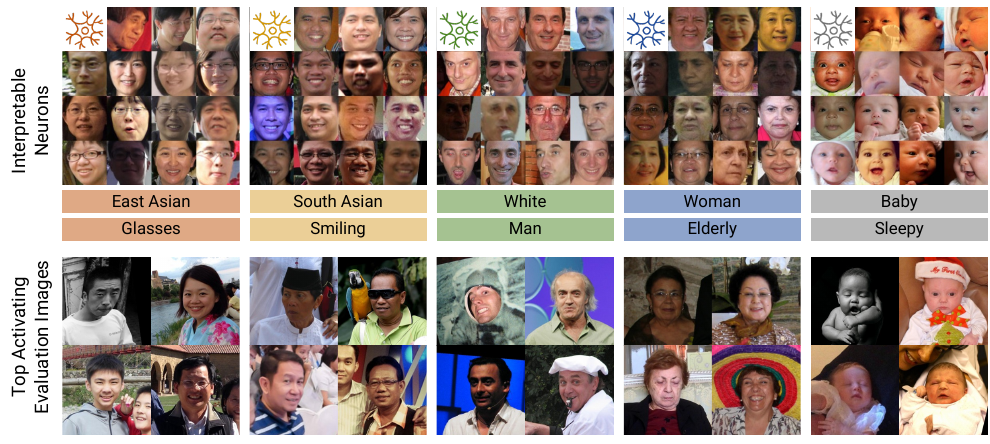}
    \caption{\textbf{Interpretable social neurons.} We visualize the top activating training (top row) and evaluation (bottom row) images for each social neuron labeled with two human-interpretable concepts.} 
    \label{fig:neurons_sup2}
\end{figure*}

The interpretability of automatically selected social neurons is further supported by illustrations in Figures~\ref{fig:neurons_sup} and \ref{fig:neurons_sup2}. While random neurons tend to activate on images containing a mixture of social demographics, the social neurons, such as those encoding gender, age, or race, exhibit selective activation patterns that correspond to specific social attributes (Figure~\ref{fig:neurons_sup}). Figure~\ref{fig:neurons_sup2} depicts top activating images for both training and evaluation images, along with the human-labeled concepts. Together, these visualizations show that the identified neurons consistently encode specific social attribute concepts.

\begin{figure*}[!t]
    \centering
    \includegraphics[width=\linewidth]{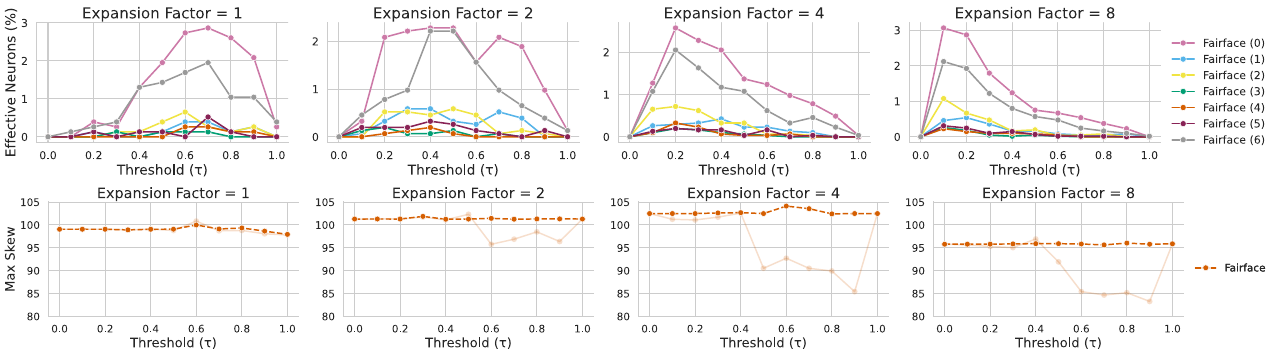}
    \caption{\textbf{Proportion of effective age neurons (top) and corresponding Max Skew scores (bottom) of CLIP (ViT-B/16) \emph{image} encoder.} The expansion factor 8 shows the lowest bias scores across thresholds (0: 3-9, 1: 10-19, 2: 20-29, 3: 30-39, 4: 40-49, 5: 50-59, 6: 60-69).}
    \label{fig:exp_factor_perf_age}
\end{figure*}

\begin{figure*}[!t]
    \centering
    \includegraphics[width=\linewidth]{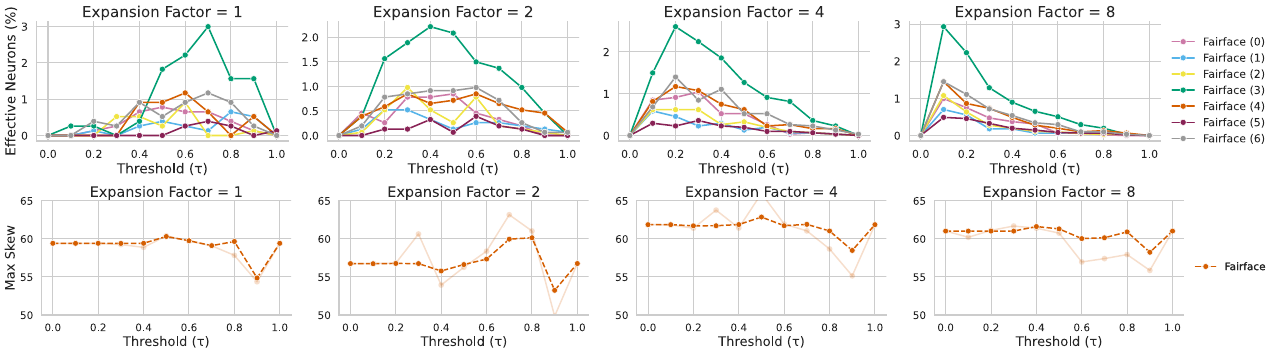}
    \caption{\textbf{Proportion of effective race neurons (top) and corresponding Max Skew scores (bottom) of CLIP (ViT-B/16) \emph{image} encoder.} The expansion factors 2 and 8 show the lowest and the most stable bias scores, respectively,  across thresholds (0: White, 1: Southeast Asian, 2: Middle Eastern, 3: Black, 4: Indian, 5: Latino Hispanic, 6: East Asian).}
    \label{fig:exp_factor_perf_race}
\end{figure*}

\begin{figure*}[!t]
    \centering
    \includegraphics[width=\linewidth]{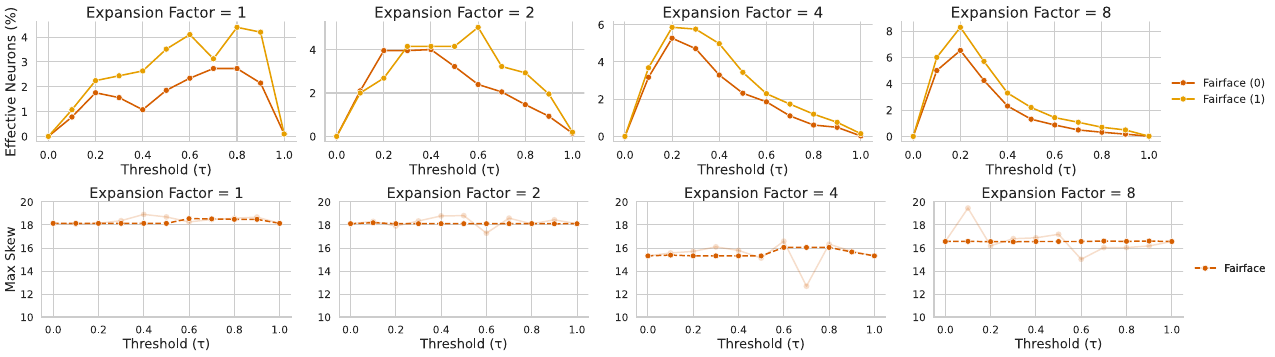}
    \caption{\textbf{Proportion of effective age neurons (top) and corresponding Max Skew scores (bottom) of CLIP (ViT-L/14@336) \emph{image} encoder.} The expansion factors 4 and 8 show the lowest bias scores across thresholds (0: Male, 1: Female).}
    \label{fig:exp_factor_perf_large}
\end{figure*}

\begin{figure*}[!t]
    \centering
    \includegraphics[width=\linewidth]{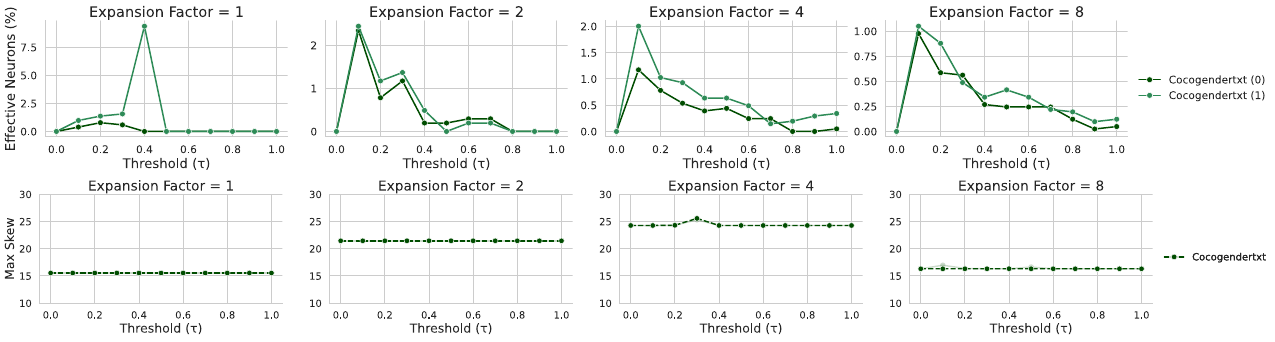}
    \caption{\textbf{Proportion of effective race neurons (top) and corresponding Max Skew scores (bottom) of CLIP (ViT-L/14@336) \emph{text} encoder.} The expansion factors 1 and 8 overall show the lowest bias scores across thresholds (0: Male, 1: Female).}
    \label{fig:exp_factor_perf_text_large}
\end{figure*}

\begin{figure*}[!t]
    \centering
    \includegraphics[width=\linewidth]{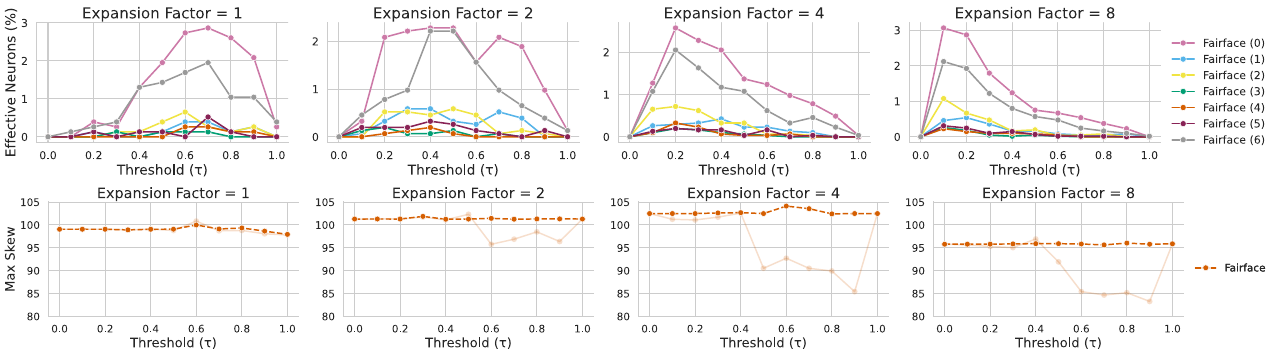}
    \caption{\textbf{Proportion of effective age neurons (top) and corresponding Max Skew scores (bottom) of CLIP (ViT-L/14@336) \emph{image} encoder.} The expansion factor 8 shows the lowest bias scores across thresholds (0: 3-9, 1: 10-19, 2: 20-29, 3: 30-39, 4: 40-49, 5: 50-59, 6: 60-69).}
    \label{fig:exp_factor_perf_large_age}
\end{figure*}

\begin{figure*}[!t]
    \centering
    \includegraphics[width=\linewidth]{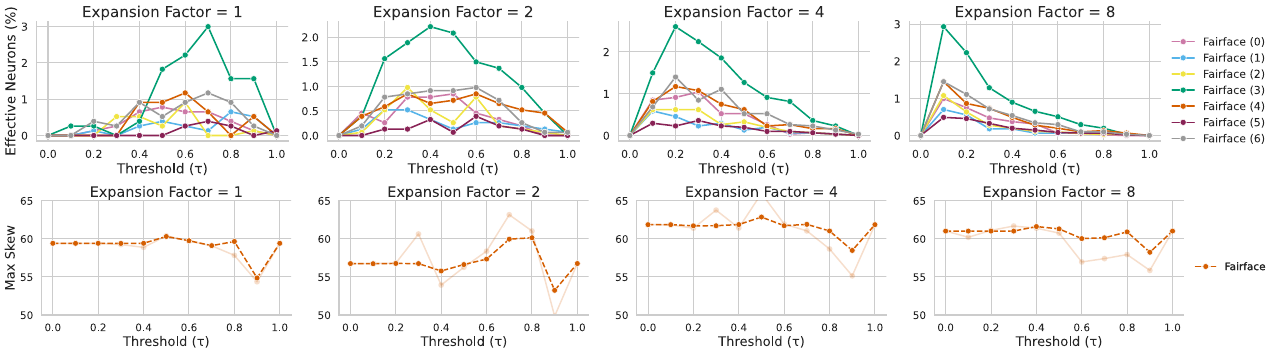}
    \caption{\textbf{Proportion of effective race neurons (top) and corresponding Max Skew scores (bottom) of CLIP (ViT-L/14@336) \emph{image} encoder.} The expansion factors 2 and 8 show the lowest and the most stable bias scores, respectively, across thresholds (0: White, 1: Southeast Asian, 2: Middle Eastern, 3: Black, 4: Indian, 5: Latino Hispanic, 6: East Asian).}
    \label{fig:exp_factor_perf_large_race}
\end{figure*}

\begin{figure*}[!t]
    \centering
    \includegraphics[width=\linewidth]{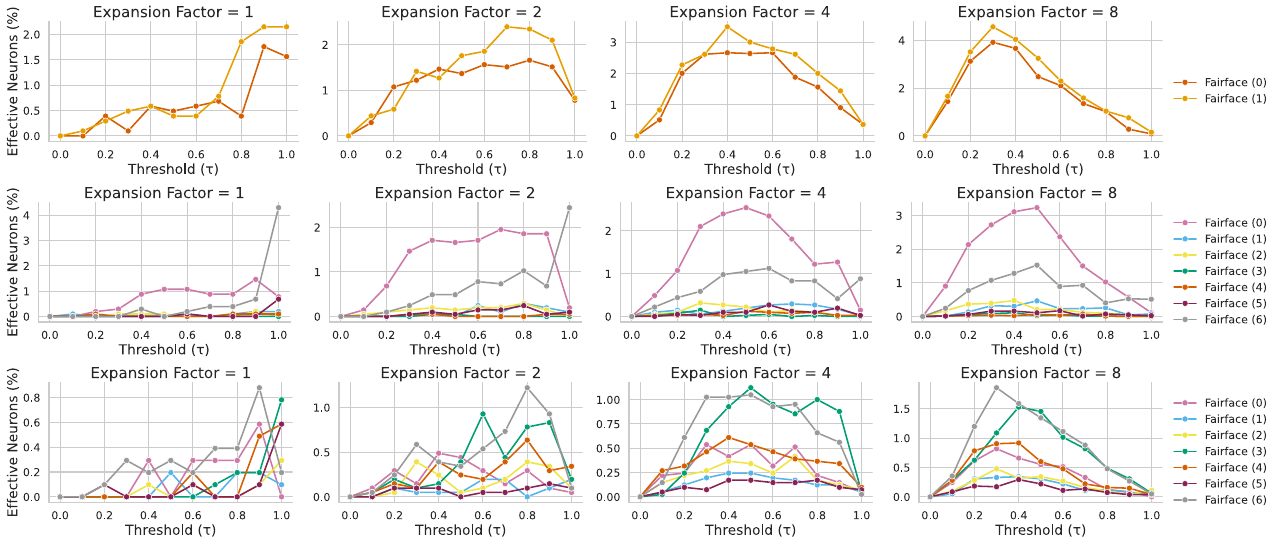}
    \caption{\textbf{Proportion of effective gender (top), age (middle), and race (bottom) neurons of InternVL2-8B image encoder.} There is a similar trend of effective neuron proportions across expansion factors for different social attributes (\emph{Gender}--0: Male, 1: Female; \emph{Age}--0: 3-9, 1: 10-19, 2: 20-29, 3: 30-39, 4: 40-49, 5: 50-59, 6: 60-69; \emph{Race}--0: White, 1: Southeast Asian, 2: Middle Eastern, 3: Black, 4: Indian, 5: Latino Hispanic, 6: East Asian).}
    \label{fig:exp_factor_intern}
\end{figure*}

\begin{figure*}[!t]
    \centering
    \includegraphics[width=\linewidth]{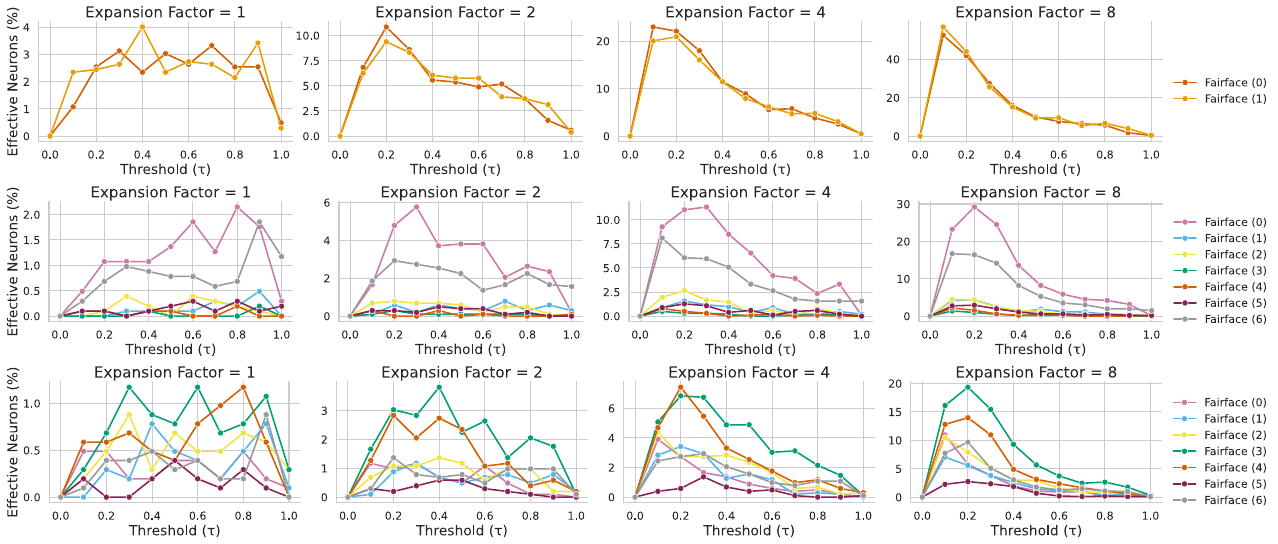}
    \caption{\textbf{Proportion of effective gender (top), age (middle), and race (bottom) neurons of LLaVA-1.5-7b-hf image encoder.} There is a similar trend of effective neuron proportions across expansion factors for different social attributes (\emph{Gender}--0: Male, 1: Female; \emph{Age}--0: 3-9, 1: 10-19, 2: 20-29, 3: 30-39, 4: 40-49, 5: 50-59, 6: 60-69; \emph{Race}--0: White, 1: Southeast Asian, 2: Middle Eastern, 3: Black, 4: Indian, 5: Latino Hispanic, 6: East Asian).}
    \label{fig:exp_factor_llava}
\end{figure*}

\begin{figure*}[!t]
    \centering
    \includegraphics[width=\linewidth]{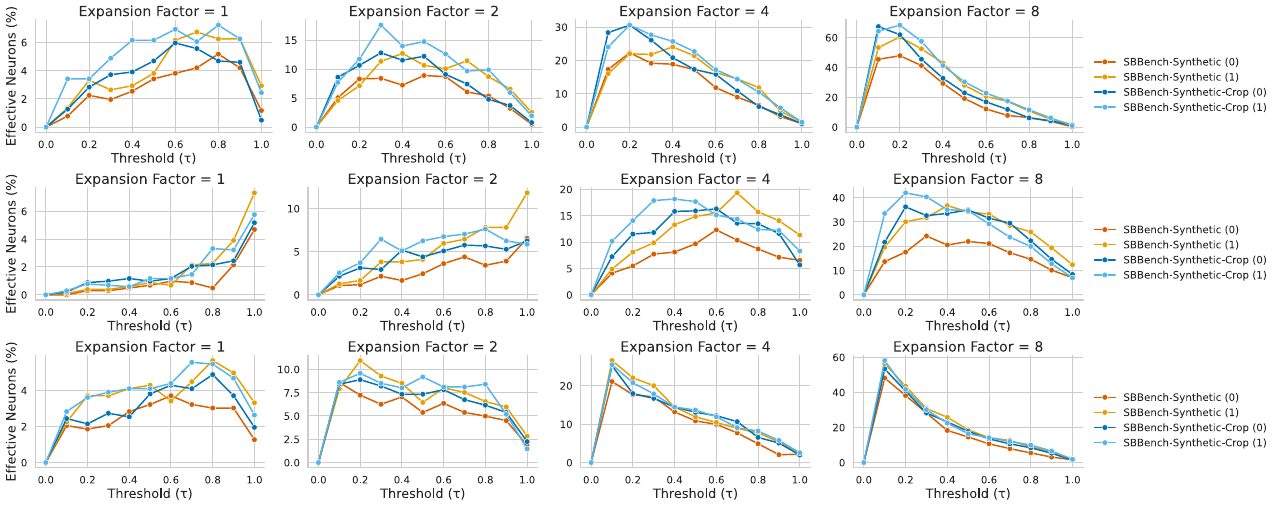}
    \caption{\textbf{Proportion of effective gender neurons of LLaVA-1.5-7b-hf (top), InternVL2-8B (middle), and LLaVAOneVision (bottom) image encoder.} There is a similar trend of effective neuron proportions across expansion factors for different models, even when trained and probed with synthetic datasets (0: Male, 1: Female).}
    \label{fig:exp_factor_sbbench_gender}
\end{figure*}

\begin{figure*}[!t]
    \centering
    \includegraphics[width=\linewidth]{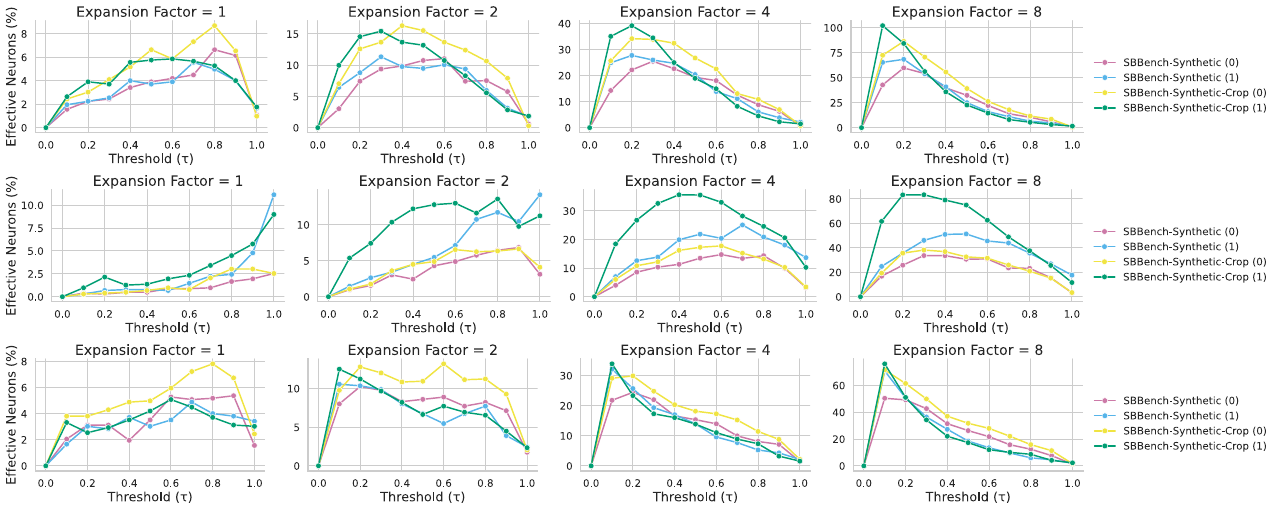}
    \caption{\textbf{Proportion of effective age neurons of LLaVA-1.5-7b-hf (top), InternVL2-8B (middle), and LLaVAOneVision (bottom) image encoder.} There is a similar trend of effective neuron proportions across expansion factors for different models, even when trained and probed with synthetic datasets (0: Old, 1: Young).}
    \label{fig:exp_factor_sbbench_age}
\end{figure*}

To validate the consistency of the robustness and optimal configuration of the neuron disentanglement, we provide results of the proportion of effective social neurons and corresponding Max Skew scores across various social attributes and models. 
Figures \ref{fig:exp_factor_perf_age} and  \ref{fig:exp_factor_perf_race} are the plots when modulating age and race neurons in the image encoder of CLIP (ViT-B/16)~\citep{radford2021learning}, attached with SAE trained using the FairFace dataset. Figures \ref{fig:exp_factor_perf_large} and \ref{fig:exp_factor_perf_text_large} are the plots when modulating gender neurons in the image and text encoder of CLIP (ViT-L/14@336). Same as CLIP (ViT-B/16), we provide the plots when modulating age and race neurons in the image encoder of CLIP (ViT-L/14@336) in Figures~\ref{fig:exp_factor_perf_large_age} and \ref{fig:exp_factor_perf_large_race}. We also show the effective social neurons when deactivating gender, age, and race neurons of InternVL2-8B~\citep{chen2024internvl} (Figure~\ref{fig:exp_factor_intern}) and LLaVA-1.5-7b-hf~\citep{liu2023visual} (Figure~\ref{fig:exp_factor_llava}). Lastly, Figures \ref{fig:exp_factor_sbbench_gender} and \ref{fig:exp_factor_sbbench_age} illustrate the proportion of effective gender and age neurons found using the image encoder of LLaVA-1.5-7b-hf, InternVL2-8b, and LLaVAOneVision~\citep{li2024llava}. All these plots reveal a similar trend of the effective neuron proportion across expansion factors and thresholds, despite the difference in probing social attributes, models, and SAE training data.

Furthermore, Table~\ref{table:rebuttal_intersection} demonstrates that our method effectively controls intersectional bias by leveraging SAE's disentanglement capacity. It selectively achieves lower MaxSkew scores (\eg, $|$Race$| < |$Age$| < |$Age $\times$ Race$| < |$ Gender $\times$ Age $\times$ Race$|$ for Age Skew).

\begin{table}[!t]
\centering
\caption{\textbf{SBBench (categories: age and gender) accuracy of \debiaslens applied to LVLM.} The best performance is achieved when SAE is trained and age neurons are selected using the FairFace datasets, measured using a rule-based and model-based evaluation.}
\begin{adjustbox}{max width=\columnwidth}
\begin{tabular}{lccccc}
\toprule

\textbf{Methods} & \textbf{Eval} & \textbf{Train Data} & \textbf{Probing Data} & \textbf{Gender} & \textbf{Age} \\
\midrule

InternVL2-8B & Rule & \tikzxmark  & \tikzxmark  & 83.83 & 43.11 \\
 \debiaslens & Rule & SB-Syn & SB-Syn  & 84.84 & 45.30 \\
 \debiaslens & Rule & SB-Syn-Crop & SB-Syn-Crop  & 84.51 & 44.13 \\
\debiaslens & Rule & FairFace & SB-Syn  & 84.64 & 45.55 \\
 \debiaslens & Rule & FairFace & SB-Syn-Crop & 84.74 & 44.77 \\
 \debiaslens & Rule & FairFace & FairFace (0.6)  & 86.60 & 47.52 \\
 \debiaslens & Rule & FairFace & FairFace (1.0) & 87.87 & 48.51 \\

\midrule

InternVL2-8B & Phi & \tikzxmark  & \tikzxmark  & 85.97 & 50.35 \\
 \debiaslens & Phi & SB-Syn & SB-Syn  & 87.20 & 51.48 \\
 \debiaslens & Phi & SB-Syn-Crop & SB-Syn-Crop  & 86.03 & 50.21 \\
 \debiaslens & Phi & FairFace & SB-Syn  & 86.91 & 51.62 \\
 \debiaslens & Phi & FairFace & SB-Syn-Crop  & 86.23 & 50.31 \\
 \debiaslens & Phi & FairFace & FairFace (0.6)  & 88.39 & 52.54 \\
 \debiaslens & Phi & FairFace & FairFace (1.0)  & 89.49 & 53.77 \\

\bottomrule
\end{tabular}
\end{adjustbox}
\label{tab:sbbench_supp}
\vspace{-1em}
\end{table}

\paragraph{Data Distribution Effects}

While the efficacy of gender neurons is detailed in the main text, we further demonstrate the impact of modulating age neurons, with results presented in Table~\ref{tab:sbbench_supp}. Similar to the main findings, social neurons found using SAE trained with the FairFace dataset seem to show the most improvement in accuracy. Also, the SAE trained and probed using cropped images from SBBench-Syn-Crop seem to show better performance with the gender neurons but not for the age neurons (Table~\ref{tab:sbbench_supp}). One of the reasons may be due to a limited amount of newly synthesized training datasets compared to FairFace (Table~\ref{table:app_groups}). Despite this, the social neurons found using the FairFace dataset can be generalized to a synthesized evaluation dataset. This suggests that our selected neurons indeed correspond to human-interpretable social attribute concepts (\eg, gender, age).

However, these neurons show lower \emph{specificity}, unlike the neurons discovered in VLMs. Concretely, modulating gender neurons seems to show better performance for questions corresponding to both gender and age categories. For instance, both the gender and age accuracies are $+1.75$ $(+1.58)$ and $+2.44$ $(+3.15)$ higher when the gender neurons are deactivated (training data: FairFace \& probing data: SB-Syn-Crop) evaluated using a rule-based approach (Phi-4~\citep{abdin2024phi}). This implies that although these social neurons are disentangled, the effect on performance may not always be localized to a single attribute, but instead propagates across intersectional demographics, especially in larger LVLMs.

\paragraph{Ablation Study}

We present detailed results of the effect of weight proportion ($\alpha$) for VLMEvalKit~\citep{duan2024vlmevalkit} and VLAGenderBias (VLA)~\citep{girrbach2025revealing} in Tables
\ref{tab:weighted_sum_sup} and \ref{tab:vla_supp}, respectively. Supporting our original claim, weighting more SAE decoded embeddings generally results in lower general performance (Table~\ref{tab:weighted_sum_sup}) and gender disproportion rate (Table~\ref{tab:vla_supp}). Furthermore, the effect on general performance shows a stronger influence when modulating \emph{all} automatically selected gender neurons (corresponding to \emph{Fairface $-$ All neurons} in Table~\ref{tab:weighted_sum_sup}), instead of the top neurons per social attribute group are selected (\ie, \debiaslens, corresponding to \emph{Fairface $-$ Top neurons}).

\begin{table}[!t]
\centering
\caption{\textbf{General performance ($\uparrow$) on varying weighted proportion for LVLMs.} The general VLM performance overall decreases as the weight proportion of the SAE decoded embedding increases.}
\resizebox{\linewidth}{!}{
\begin{tabular}{lcccc}
\toprule
\textbf{$\alpha$} & \textbf{Percep}~\citep{chaoyou2023mme} & \textbf{Reason}~\citep{chaoyou2023mme} & \textbf{MMMU}~\citep{yue2024mmmu} & \textbf{Seed2}~\citep{li2024seed} \\

\midrule
\multicolumn{5}{c}{LLaVA-1.5-7b-hf~\cite{liu2023visual} (\emph{Fairface $-$ Top neurons})} \\
\midrule
 0.0 & 1205.10 & 235.00 & 0.30 & 0.59 \\

 0.2 & 1252.50 & 226.78 & 0.29 & 0.59 \\

 0.4 & 1240.75 & 255.35 & 0.29 & 0.58 \\

 0.5 & 1209.25 & 274.64 & 0.30 & 0.58 \\

 0.6 & 1187.84 & 266.42 & 0.30 & 0.57 \\

 0.8 & 1096.65 & 263.57 & 0.31 & 0.56 \\

 1.0 & 930.55 & 221.78 & 0.22 & 0.53 \\
\midrule
 \multicolumn{5}{c}{InternVL2-8b~\cite{chen2024internvl} (\emph{Fairface $-$ Top neurons})} \\
\midrule
 0.0 & 1646.79 & 536.78 & 0.43 & 0.75 \\

 0.2 & 1657.39 & 526.78 & 0.43 & 0.75 \\

 0.4 & 1644.58 & 525.00 & 0.44 & 0.75 \\

 0.5 & 1618.70 & 492.85 & 0.40 & 0.75 \\

 0.6 & 1616.65 & 478.21 & 0.39 & 0.74 \\

 0.8 & 1603.35 & 445.35 & 0.43 & 0.73 \\

 1.0 & 1561.92 & 401.07 & 0.44 & 0.72 \\
 \midrule
\multicolumn{5}{c}{InternVL2-8b~\cite{chen2024internvl} (\emph{Fairface $-$ All neurons})} \\
\midrule
 0.0 & 1646.79 & 536.78 & 0.43 & 0.75 \\

 0.2 & 1663.89 & 529.28 & 0.39 & 0.75 \\

 0.4 & 1643.56 & 527.14 & 0.44 & 0.75 \\

 0.5 & 1622.05 & 492.85 & 0.40 & 0.74 \\

 0.6 & 1609.96 & 477.85 & 0.39 & 0.74 \\

 0.8 & 1592.31 & 452.14 & 0.44  & 0.73 \\

1.0 & 1549.64 & 381.07 & 0.39 & 0.72 \\
 \midrule
\multicolumn{5}{c}{InternVL2-8b~\cite{chen2024internvl} (\emph{Fairface $-$ All neurons $-$ Negative activations})} \\
\midrule
 0.0 & 1646.79 & 536.78 & 0.43 & 0.75 \\

 0.2 & 524.09 & 223.57 & 0.33 & 0.38 \\

 0.4 & 524.09 & 223.57 & 0.33 & 0.38 \\

 0.5 & 524.09 & 223.57 & 0.33 & 0.38 \\

 0.6 & 524.09 & 223.57 & 0.33 & 0.38 \\

 0.8 & 524.09 & 223.57 & 0.33 & 0.38 \\

 1.0 & 524.84 & 223.57 & 0.33 & 0.38 \\
\bottomrule
\end{tabular}
}
\label{tab:weighted_sum_sup}
\end{table}
\begin{table}[!t]
\centering
\caption{\textbf{Gender disproportion rate ($\downarrow$) across varying weighted proportions for LLaVA-1.5-7b-hf.} The disproportion rate decreases as the weight proportion of the SAE decoded embedding increases.}
\resizebox{\linewidth}{!}{
\begin{tabular}{lcccc}
\toprule
\textbf{$\alpha$} & \textbf{Occupations} & \textbf{Trait} & \textbf{Trait (gendered)} & \textbf{Skills} \\
\midrule
0.0 & 0.3500 & 0.7000 & 0.7500 & 0.7143 \\
0.2 & 0.3250 & 0.6000 & 0.7083 & 0.6667 \\
0.4 & 0.3250 & 0.6000 & 0.5833 & 0.6190 \\
0.5 & 0.3250 & 0.5500 & 0.5417 & 0.6190 \\
0.6 & 0.3250 & 0.5500 & 0.5417 & 0.6190 \\
0.8 & 0.2750 & 0.5500 & 0.5417 & 0.5714 \\
1.0 & 0.2250 & 0.5500 & 0.5833 & 0.2857 \\
\bottomrule
\end{tabular}
}
\label{tab:vla_supp}
\vspace{-1em}
\end{table}

\begin{table}[!t]
\centering
\caption{\textbf{Social Attribute Predictability Results.}}
\resizebox{\linewidth}{!}{
\begin{tabular}{lcccc}
\toprule 
Representation & $\alpha$ & Gender Acc & Age Acc & Race Acc\\
\midrule
$\mathbf{v}$ (original) & 0.0 & 95.9 & 55.6 & 71.0 \\
$\mathbf{v'}$ (mixed) & 0.6 & 95.2 & 56.2 & 70.8 \\
$\hat{\mathbf{v}}$ (SAE recon) & 1.0 & 92.7 & 51.4 & 62.6 \\
\bottomrule
\end{tabular}
}
\label{table:rebuttal_attr_pred}
\vspace{-1em}
\end{table}

These trends reflect the underlying trade-off in the construction of the representation $\mathbf{v}'$, which interpolates between the original feature $\mathbf{v}$ and the SAE-decoded reconstruction $\hat{\mathbf{v}}$. To better understand this trade-off, we further examine the bias properties of $\hat{\mathbf{v}}$ through both empirical and theoretical analyses.

Empirically, $\hat{\mathbf{v}}$ exhibits lower attribute predictability than both $\mathbf{v}$ and $\mathbf{v}'$ (Tab.~\ref{table:rebuttal_attr_pred}). This suggests that although biased signals may still persist in the decoded reconstructions, they are no longer concentrated in fixed latent dimensions; instead, they emerge from different subsets of active latents across samples.

Theoretically, let the SAE encoder produce sparse activations
\begin{equation}
\mathbf{z} = \sigma(\mathbf{W}_e \mathbf{v} + \mathbf{b}_e),
\end{equation}
where $\sigma(\cdot)$ is a sparsity-inducing nonlinearity (\eg, ReLU or Top-$k$).  
We define the \emph{active set} as
\begin{equation}
\mathcal{A}(\mathbf{v})
=
\{ i \mid (\mathbf{W}_e \mathbf{v} + \mathbf{b}_e)_i > 0 \},
\end{equation}
namely, the indices of latent neurons activated by input $\mathbf{v}$.  
Let $\mathbf{D}_{\mathcal{A}(\mathbf{v})}$ denote the diagonal masking matrix whose $(i,i)$-th entry equals $1$ if $i \in \mathcal{A}(\mathbf{v})$ and $0$ otherwise.
The SAE reconstruction can then be written as
\begin{equation}
\hat{\mathbf{v}}
=
\mathbf{W}_d \mathbf{D}_{\mathcal{A}(\mathbf{v})} \mathbf{W}_e \mathbf{v}
+
\mathbf{c}_{\mathcal{A}(\mathbf{v})},
\end{equation}
where $\mathbf{c}_{\mathcal{A}(\mathbf{v})}$ absorbs bias terms. For a fixed active set, the mapping is linear; however, since $\mathcal{A}(\mathbf{v})$ varies with the input, the overall function is piecewise linear and globally non-linear. The effective linear transformation
\begin{equation}
\mathbf{M}_{\mathcal{A}(\mathbf{v})}
=
\mathbf{W}_d \mathbf{D}_{\mathcal{A}(\mathbf{v})} \mathbf{W}_e
\end{equation}
therefore changes across inputs.

A single global separating direction $\mathbf{w}$ would require
\begin{equation}
\mathbf{w}^\top \mathbf{M}_{\mathcal{A}_1}
=
\mathbf{w}^\top \mathbf{M}_{\mathcal{A}_2}
\quad
\forall \mathcal{A}_1, \mathcal{A}_2,
\end{equation}
which holds only in degenerate cases, such as \emph{constant active sets}, \ie,
\begin{equation}
\mathcal{A}(\mathbf{v}_1) = \mathcal{A}(\mathbf{v}_2)
\quad \forall \mathbf{v}_1, \mathbf{v}_2.
\end{equation}
In that case, $\mathbf{D}_{\mathcal{A}(\mathbf{v})}$ becomes a fixed matrix, and the mapping reduces to a single global linear transformation. However, under typical sparse activation regimes, different inputs induce different active sets (other than the common active sets mapping to our selected social neurons). Hence, no stable global linear direction can consistently separate social attributes in $\hat{\mathbf{v}}$.

Together, these findings explain why increasing the weight on $\hat{\mathbf{v}}$ systematically reduces measurable linear bias through disrupting globally aligned attribute directions while introducing a controllable drop in general performance.

\section{Limitation and Future Work}\label{app:limit_future}

While \debiaslens presents a transparent and effective approach for identifying and mitigating bias through monosemantic social neurons, several limitations remain that open important directions for future research.
First, our method relies on the quality and coverage of the existing SAE training data. Although our experimental results demonstrate that the FairFace dataset is sufficient for finding social neurons, they may underrepresent more subtle or culturally specific forms of bias. This could potentially limit the granularity of the social neurons. Also, the existing dataset does not currently include fine-grained social attribute labels, which limits its ability to account for more inclusive and diverse demographic representations. Hence, we urge future works to collect large-scale, demographically balanced, and globally diverse facial datasets that encompass overlooked or underrepresented populations, enabling more robust and inclusive debiasing.

Second, our intervention currently assumes that social attributes can be cleanly disentangled within a small set of neurons. While this assumption held empirically, complex or intersectional biases (\eg, age × gender × race) may require more nuanced structures such as hierarchical or multi-branch SAEs.
Lastly, we focus on neuron-level modulation and do not explicitly examine how higher-level model components, such as image-text alignments, interact with these social neurons. We leave as future work to conduct systematic interventions that adjust not only neuron activations but also the pathways through which bias propagates.
We hope that \debiaslens inspires future research toward building fair, transparent, and socially responsible VLMs and LVLMs.

\clearpage
\newpage

\begin{table*}[!t]
\centering
\caption{\textbf{Statistics of training and evaluation data.} The table presents statistics for group labels for each social attribute across the datasets used in this work. Note that every image includes one human or face, except for the SBBench evaluation dataset, which includes two humans per image (female/male and young/old for the gender and age categories). }
\begin{adjustbox}{max width=0.85\linewidth}
\begin{tabular}{p{5cm} ccc p{6cm}}
\toprule
\textbf{Dataset}       & \textbf{Train} & \textbf{Eval} & \textbf{Data size} & \textbf{Social bias attributes} \\
\midrule

\multirow{22}{*}{FairFace} 
& \multirow{22}{*}{\checkmark} 
& 
& \multirow{22}{*}{86{,}744} 
& \begin{itemize}[leftmargin=*]
    \item \textbf{Gender}: Male (53\%), Female (47\%)
\end{itemize} \\
& & & & \begin{itemize}[leftmargin=*]
    \item \textbf{Age}:
    \begin{itemize}[leftmargin=1em]
        \item 0--2 (2\%)
        \item 3--9 (12\%)
        \item 10--19 (11\%)
        \item 20--29 (30\%)
        \item 30--39 (22\%)
        \item 40--49 (12\%)
        \item 50--59 (7\%)
        \item 60--69 (3\%)
        \item $>$70 (1\%)
    \end{itemize}
\end{itemize} \\
& & & & \begin{itemize}[leftmargin=*]
    \item \textbf{Race}:
    \begin{itemize}[leftmargin=1em]
        \item White (19\%)
        \item Latino Hispanic (15\%)
        \item Indian (14\%)
        \item East Asian (14\%)
        \item Black (14\%)
        \item Southeast Asian (12\%)
        \item Middle East Asian (11\%)
    \end{itemize}
\end{itemize} \\
& & & & \\[-1em] 
\midrule

Cocogender (\& Cocogendertxt)   & \multirow{1}{*}{\checkmark}        &            & \multirow{1}{*}{12,454}     & \textbf{Gender}: Male (65\%), Female (35\%)                 \\
\midrule
\multirow{1}{*}{CelebA}        & \checkmark        &            & \multirow{1}{*}{202,599} & \textbf{Gender}: Male (43\%), Female (58\%)                \\
\midrule
\multirow{1}{*}{Bias in Bios}  & \multirow{1}{*}{\checkmark}        &            & \multirow{1}{*}{257,478}    & \textbf{Gender}: Male (54\%), Female (46\%)               \\
\midrule
\multirow{2}{*}{SBBench-Syn}        & \multirow{1}{*}{\checkmark} &           & \multirow{1}{*}{206} & \textbf{Gender}: Male (53\%), Female (47\%)     \\ 
\cmidrule{4-5}
     &  &           &  \multirow{1}{*}{333} & \textbf{Age}: Old (74\%), Young (26\%)    \\   
\midrule
     
\multirow{2}{*}{SBBench-Syn-Crop}         & \multirow{2}{*}{\checkmark}  &           &   \multirow{1}{*}{406} & \textbf{Gender}: Male (45\%), Female (55\%)       \\   
\cmidrule{4-5}
      &  &           & \multirow{1}{*}{511} & \textbf{Age}: Old (69\%), Young (31\%)     \\   
      
\midrule

\multirow{15}{*}{FairFace} 
&
& \multirow{15}{*}{\checkmark}  
& \multirow{15}{*}{10{,}324} 
& \begin{itemize}[leftmargin=*]
    \item \textbf{Gender}: Male (50\%), Female (50\%)
\end{itemize} \\
& & & & \begin{itemize}[leftmargin=*]
    \item \textbf{Age}:
    \begin{itemize}[leftmargin=1em]
        \item 0--2 (1\%)
        \item 3--9 (12\%)
        \item 10--19 (11\%)
        \item 20--29 (31\%)
        \item 30--39 (21\%)
        \item 40--49 (12\%)
        \item 50--59 (7\%)
        \item 60--69 (3\%)
        \item $>$70 (1\%)
    \end{itemize}
\end{itemize} \\
& & & & \begin{itemize}[leftmargin=*]
    \item \textbf{Race}: same ratios as training data \end{itemize} \\
& & & & \\[-1em] 

\midrule
\multirow{1}{*}{VLAGenderBias} &  & \multirow{1}{*}{\checkmark}          &  \multirow{1}{*}{5,000}  & \textbf{Gender}: Male (50\%), Female (50\%) \\
\midrule
\multirow{2}{*}{SBBench}     &          & \multirow{2}{*}{\checkmark}           & \multirow{1}{*}{3,094} & \textbf{Gender}: Male (50\%), Female (50\%)  \\   
      &          &          & 2,838 & \textbf{Age}: Old (50\%), Young (50\%)   \\   
\bottomrule
\end{tabular}
\end{adjustbox}
\label{table:app_groups}
\end{table*}

\end{document}